\documentclass[12pt]{article}
\usepackage{fullpage,graphicx,amsmath,amssymb,verbatim,xcolor}
\usepackage[small,bf]{caption}

\usepackage[noend]{algpseudocode}
\usepackage{algorithm}
\usepackage{url}
\usepackage{tikz}
\usetikzlibrary{intersections}

\usepackage{subcaption}

\usetikzlibrary{positioning,calc,shapes,arrows}
\tikzstyle{block} = [rectangle, draw,
text width=7em, text centered, minimum height=5em]
\tikzstyle{line} = [draw, -latex',line width=1mm]

\usepackage{hyperref}
\hypersetup{
colorlinks   = true,    
urlcolor     = blue,    
linkcolor    = blue,    
citecolor    = red      
}

\usepackage{cleveref}
\crefname{equation}{}{}
\Crefname{equation}{}{}

\newcommand{\ones}{\mathbf 1}

\newcommand{\reals}{{\mbox{\bf R}}}

\newcommand{\BEAS}{\begin{eqnarray*}}
\newcommand{\EEAS}{\end{eqnarray*}}
\newcommand{\BEA}{\begin{eqnarray}}
\newcommand{\EEA}{\end{eqnarray}}
\newcommand{\BEQ}{\begin{equation}}
\newcommand{\EEQ}{\end{equation}}
\newcommand{\BIT}{\begin{itemize}}
\newcommand{\EIT}{\end{itemize}}

\newcommand{\diag}{\mathop{\bf diag}}
\newcommand{\blkdiag}{\mathop{\bf blkdiag}}

\newcommand{\argmin}{\mathop{\rm argmin}}
\newcommand{\argmax}{\mathop{\rm argmax}}

\newcommand{\sign}{\mathop{\bf sign}}

\newcommand{\eg}{{\it e.g.}}
\newcommand{\ie}{{\it i.e.}}

\newcounter{algorithmctr}[section]
\renewcommand{\thealgorithmctr}{\thesection.\arabic{algorithmctr}}
\newenvironment{algdesc}%
   {\refstepcounter{algorithmctr}
   \begin{list}{}{%
       \setlength{\rightmargin}{0\linewidth}%
       \setlength{\leftmargin}{.05\linewidth}}%
       \rmfamily\small
       \item[]{\setlength{\parskip}{0ex}\hrulefill\par%
        \nopagebreak{\bfseries\textsf{Algorithm \thealgorithmctr~}}}}%
   {{\setlength{\parskip}{-1ex}\nopagebreak\par\hrulefill} 
   \end{list}}

   {\refstepcounter{algorithmctr}
   \begin{list}{}{%
       \setlength{\rightmargin}{0\linewidth}%
       \setlength{\leftmargin}{.05\linewidth}}%
       \rmfamily\small
       \item[]{\setlength{\parskip}{0ex}\hrulefill\par%
        \nopagebreak{\bfseries\textsf{Algorithm}}}}%
   {{\setlength{\parskip}{-1ex}\nopagebreak\par\hrulefill} 
   \end{list}}

\title{Factor Fitting, Rank Allocation, and Partitioning\\ in Multilevel Low Rank Matrices}

\author{Tetiana Parshakova \and Trevor Hastie \and Eric Darve \and Stephen Boyd}

\date{}

\begin{document}
\maketitle

\begin{abstract}
We consider multilevel low rank (MLR) matrices, defined as
a row and column permutation of
a sum of matrices, each one a block diagonal refinement of the previous one, 
with all blocks low rank given in factored form.
MLR matrices extend low rank matrices
but share many of their properties, such as the total storage required and
complexity of matrix-vector multiplication.
We address three problems that arise in fitting a given matrix by an 
MLR matrix in the Frobenius norm.  The first problem is factor fitting,
where we adjust the factors of the MLR matrix.
The second is rank allocation, where we choose the ranks of the blocks 
in each level, subject to the total rank having a given value, which preserves
the total storage needed for the MLR matrix.
The final problem is to choose the hierarchical partition of rows and 
columns, along with the ranks and factors.
This paper is accompanied by an open source package that implements the
proposed methods. 
\end{abstract}

\clearpage
\tableofcontents
\clearpage

\section{Introduction}
Multilevel low rank (MLR) matrices are an extension of low rank matrices
and factor models of covariance matrices.
They generalize low rank matrices in the sense of giving a substantial
reduction in storage and speed up in computing matrix-vector products,
compared to a full dense matrix.
MLR matrices are closely related to hierarchical matrices,
which have been studied since at least the late 1980s.
There are different definitions of hierarchical matrices, and our
MLR notion is close to, but not the same as, others that have
been proposed.
Our focus is on three problems associated with 
fitting a given full matrix with an MLR matrix with respect 
to the Frobenius norm.

The first problem we address is factor fitting, which involves finding 
the factors of the MLR matrix to best fit a given matrix in Frobenius norm.
We present two complementary block coordinate descent algorithms for this 
fitting problem.
One alternates between updating the left and the right factors,
each of these done by solving a structured least squares problem.
The other method cycles over optimizing the matrices at each level of 
the hierarchy, which can be done via the singular value decomposition (SVD).

The second problem is rank allocation, where we divide up a fixed total rank 
across the levels of the hierarchy, which preserves the storage required.
We propose a heuristic based on the SVD of the matrices at each level 
of the hierarchy, which estimates the change in fitting objective when
the rank allocated to each level is either incremented or decremented,
and then chooses a candidate exchange of rank across the levels.

The third problem is general MLR fitting, where only the 
matrix to be approximated and a total rank are given.
This problem involves finding the hierarchical partitioning of rows and columns, 
as well as rank allocation and fitting the factors.
We propose a simple nested dissection of the rows and columns using spectral
bi-clustering.

Except for very simple cases (such as a low rank matrix)
the methods we propose are
heuristic, and can and do converge to non-global local optima.
The methods we propose are related to prior work 
(as described below in more detail),
but we believe that our methods for factor fitting and rank allocation
are novel.

\subsection{Prior work}

\paragraph{Low rank approximation.}
The problem of finding a low rank approximation to a given matrix 
is readily solved via the singular value decomposition (SVD)
\cite{datta2010numerical, darve2021numerical}.
The solution was originally proposed in 1907~\cite{Schmidt1907} in the context 
of integral operators and rediscovered for matrices in 
1936~\cite{eckart1936approximation}.
Since then the problem has been extensively studied~\cite{kishore2017literature}. 
Numerous algorithms have been devised to solve the problem approximately 
with a reduced computational cost, 
including QR decomposition with pivoting~\cite{stewart1998matrix}, rank 
revealing QR factorization~\cite{CHAN198767},
randomized algorithms~\cite{achlioptas2003database, deshpande2006adaptive} 
and cross/skeleton decomposition~\cite{bebendorf2011adaptive}.
By storing the left and right factors, low rank matrices require 
far less storage than a dense matrix with the same dimensions.
Various operations can be carried out more efficiently, \eg, 
matrix-vector multiply.

\paragraph{Hierarchical matrices.}
Hierarchical matrices ($\mathcal{H}$-matrices) have been around since at 
least the late 1980s~\cite{greengard1987fast, tyrtyshnikov1996mosaic, 
hackbusch1999sparse}, with multiple books dedicated to 
them~\cite{bebendorf2008hierarchical, hackbusch2015hierarchical}.
These matrices arise naturally from numerical solutions to integral
equations~\cite{gohberg1985linear, rokhlin1985rapid, greengard1987fast}, 
and discretization of kernel functions and finite element 
matrices~\cite[\S 10,~11]{hackbusch2015hierarchical}.

Various specific forms for hierarchical matrices have been proposed previously,
among which
the $\mathcal{H}$-matrix is the most general one~\cite{hackbusch1999sparse, 
borm2003introduction, grasedyck2003construction, 
bebendorf2008hierarchical, hackbusch2015hierarchical}.
A special case of $\mathcal{H}$-matrix is the hierarchically off-diagonal 
low-rank (HODLR) matrix~\cite{aminfar2016fast},
where the off-diagonal blocks on multiple levels are low rank.
$\mathcal{H}^2$-matrices~\cite{Hackbusch2002data, borm2003introduction} 
are a subcategory of $\mathcal{H}$-matrices with nested bases, and 
hierarchical semiseparable (HSS) 
matrices~\cite{chandrasekaran2006fast,xia2010fast,
ambikasaran2013mathcal} are a special case of HODLR matrices with nested bases. 
The fast multipole method (FMM)~\cite{greengard1987fast, 
darve2000fast, nishimura2002fast, ying2004kernel}, which is considered one 
of the top $10$ algorithms of 20th century~\cite{cipra2000best}, 
can be interpreted as relying on $\mathcal{H}^2$-matrices.

All of these types of hierarchical matrices provide storage reduction 
and enable fast matrix-vector multiplication, just like low rank matrices.
Approximate LU factorization, and solving linear equations, can be
carried out much faster than for a dense matrix of the 
same dimensions ~\cite[\S 7]{hackbusch2015hierarchical}.
This is especially useful since
many matrices that arise in scientific and engineering 
contexts are, or can be approximated by,
hierarchical matrices~\cite{kalman1963mathematical, bonnet1999boundary, 
buhmann2003radial, hsiao2006boundary}.

Direct solvers~\cite{amestoy2001mumps, henon2002pastix, chen2008algo} become 
computationally impractical for large-scale dense linear systems. Hence, iterative 
methods such as conjugate gradient~\cite{hestenes1952methods}, 
MINRES~\cite{paige1975solution}, or GMRES~\cite{saad1986gmres} should be considered.
In practice, however, iterative methods need to be accompanied by 
preconditioners~\cite{wathen_2015}. Efficient preconditioners can be developed using
fast approximate hierarchical matrix factorization.
These hierarchical preconditioners include 
$\mathcal{H}$-matrices~\cite{hackbusch1999sparse, grasedyck2003construction, 
schmitz2014fast, grasedyck2008performance},
HODLR~\cite{kong2011adaptive, aminfar2016fast, pouransari2017fast},
HSS~\cite{chandrasekaran2006fast, chandrasekaran2007fast, martinsson2009fast, 
xia2010superfast, gillman2012direct, xia2013rand} and 
others~\cite{coulier2017inverse, klockiewicz2020sparse, klockiewicz2022second}.

While we also consider hierarchical diagonal block structure as in HODLR,  
our MLR matrices are a bit different from HODLR.  We decompose our matrix
into a sum of block diagonal matrices for each level of the hierarchy with 
each block having a low rank. 
When representing MLR using HODLR format, an extra logarithmic factor 
(in the size of the matrix) is introduced in the storage.
Similarly, an extra logarithmic factor appears when 
representing HODLR using MLR format.
We present a comparison between MLR and HODLR in \S\ref{s-examples}.

\paragraph{Block low rank matrices.}
Block low rank (BLR) matrix format~\cite{amestoy2015improving} is a flat 
non-hierarchical block matrix structure. Specifically, the matrix is partitioned 
into blocks of the same size with dense diagonal blocks and low rank 
off-diagonal blocks. 
BLR matrix format has been successfully employed to accelerate sparse 
direct methods~\cite{pichon2018sparse, amestoy2019performance}, 
multifrontal method~\cite{amestoy2015improving, mary2017block},
Cholesky factorization \cite{akbudak2017tile, cao2020extreme}, 
solving boundary integral equations~\cite{al2020solving} and many 
more~\cite{weisbecker2013improving, sterling2017high}. Although LU 
factorization using BLR matrices has shown improvements, 
it is still less efficient in terms of both memory and operations compared 
to the ones based on $\mathcal{H}$-matrices \cite{amestoy2015improving}.

Our MLR matrices have similarities with BLR matrices, but they are
not the same.
Specifically, blocks in an MLR matrix have contributions from
all levels in the hierarchy, and therefore need not be low rank.

\paragraph{Butterfly matrices.}
Butterfly matrices~\cite{parker1995random} are a structured set of matrices 
that encode the recursive divide and conquer fast Fourier transform algorithm~\cite{cooley1965algorithm}.
Butterfly matrix parameterization~\cite{dao2019learning}
is motivated by the fact that  
matrices with fast matrix-vector multiplication can be characterized
as being factorizable
into products of sparse matrices~\cite{de2018two}. 
This parameterization captures many  
structured linear transformations with nearly optimal space 
and time complexity~\cite{dao2020kaleidoscope}.
It has been successfully employed for
sparse neural network training~\cite{dao2021pixelated}, specifically using
Monarch matrices~\cite{dao2022monarch}, 
which generalize butterfly matrices while being hardware efficient.

MLR and Monarch matrices share certain similarities, yet they are distinct.
Monarch matrices are parameterized as 
products of two block diagonal matrices up to permutation.
MLR, on the other hand, is a row and column permutation of a
sum of products of two block diagonal matrices.
The permutation in Monarch matrices is fixed, turning the first
block diagonal factor into a block matrix with diagonal blocks, 
while the permutation in MLR is general and is applied 
to factors in a different order.
We compare MLR and Monarch matrices empirically in \S\ref{s-examples}.

\paragraph{Hierarchical principal component analysis.}
Principal component analysis (PCA) is a technique that reduces the dimensionality 
of a set of observations while preserving as much variance as possible. 
PCA was presented back in 
1901~\cite{pearson1901liii} and since then there have been many 
developments~\cite{jolliffe2016principal}, \eg,
generalized low rank models~\cite{udell2016generalized}
which impose additional structure on the factors.
Another development is so-called multi-block component models, 
which are used when 
the data matrix is collected in blocks~\cite{smilde2003framework}. 

Hierarchical PCA (HPCA) is a type of 
multiblock PCA method~\cite{wold1996hierarchical}, which is used to increase the 
interpretability of PC scores and loadings when the number of features is large. 
HPCA is easier to interpret because the model coefficients are 
divided into multiple levels, modeling the relationship 
between and within the blocks. 
To the best of our knowledge, HPCA is related to
MLR matrices only by the fact that both models are hierarchical, \ie, 
they use multi-level hierarchical partitioning, and, as with all
the matrices discussed above, involve low rank approximation of 
blocks.

\paragraph{Factor models.} 
Factor analysis is a method for modeling observed features in terms of a smaller
number of unobserved or latent factors. It was pioneered in psychometrics in 
1904~\cite{10.2307/1412107}. 
Factor models represent a covariance matrix as a sum of a low rank matrix 
(related to the factors),
plus a diagonal matrix (sometimes called the idiosyncratic variance),
and are a special case of our definition of MLR matrix.
There are many methods for fitting factor 
(\ie, low rank plus diagonal) matrices; 
see, \eg,~\cite[\S 2, App.~A]{harman1976modern, coughlin2013analysis}.
Related concepts also include higher-order and hierarchical factor models 
~\cite{yung1999relationship, moench2013dynamic} as well as multilevel
factor models~\cite{goldstein2014multilevel, choi2018multilevel}. 
With these structures, we can characterize variations both
between and within blocks, which increases the 
interpretability of factors.

\paragraph{Spectral clustering.}  
Our method for finding a suitable hierarchical partition of the rows 
and columns relies on the well-known idea of spectral clustering.
There are various clustering techniques including connectivity-based, 
centroid-based, distribution-based, data 
stream-based and spectral clustering~\cite{xu2015comprehensive}.
Spectral methods~\cite{verma2003comparison}, in particular, 
have been very popular because of their simple and 
efficient implementation that relies on linear algebra 
packages~\cite{von2007tutorial}.
Spectral graph partitioning was first introduced in the 
1970s~\cite{hall1970r, donath1973lower, fiedler1973algebraic} 
and gained popularity in 
the 1990s~\cite{pothen1990partitioning}.

Bi-clustering (or co-clustering) is a clustering technique for simultaneous 
clustering the rows and columns
of a matrix. It was introduced in 1972~\cite{hartigan1972direct} and later 
generalized in 2000~\cite{cheng2000biclustering} 
for simultaneous clustering of genes and conditions. 
Spectral methods have proved effective in bipartite graph 
partitioning~\cite{dhillon2001co, zha2001bipartite} 
and have been successfully applied to hierarchical 
matrices as well~\cite[\S 1.4.1]{bebendorf2008hierarchical}.

\paragraph{Rank allocation.}
There are methods for fitting all of the matrix forms described above, 
including choosing the rank of the submatrices.
Previous studies have explored an adaptive rank determination
algorithm~\cite[\S 6.6]{hackbusch2015hierarchical} which sets the rank 
of each block based on a prescribed upper bound on the truncation error.
Similarly, in~\cite{massei2022hierarchical} the blocks are recursively 
partitioned until a small truncation error is achieved given 
a maximum rank per block. 
We are not aware of prior work that uses a fixed total rank across
the levels, and then adjusts the allocation of rank.
To the best of our knowledge our rank allocation
algorithm is a novel approach.

\subsection{Our contributions}
The main contributions of this paper are the following:
\begin{enumerate}
\item We introduce a novel definition of multilevel low rank matrix that, 
while closely related, differs from the traditional definitions of hierarchical
matrix.
\item We present two complementary block coordinate descent algorithms 
for factor fitting.
\item We present an algorithm for rank allocation that is able to 
re-allocate the rank assigned to each level in the hierarchy, to
improve the fitting.
\end{enumerate}
We provide an open-source package that implements these methods,
available at \begin{quote}
\url{https://github.com/cvxgrp/mlr_fitting}.
\end{quote}
We also provide a number of examples that illustrate our methods.

\subsection{Outline}

In \S\ref{s-MLR} we describe our definition of MLR matrices,
which is very close to, but not the same as, traditional
hierarchical matrices.
We describe the MLR fitting problem in \S\ref{s-fitting}.
In \S\ref{s-fitting-methods} we give two different block coordinate
descent method methods for factor fitting, where we choose the coefficients
in the MLR matrix.
In \S\ref{s-rank-alloc} we describe our rank allocation algorithm,
where a given total rank is to be allocated across the different levels
of the MLR matrix.
We describe a method for the full MLR fitting problem in 
\S\ref{s-MLR-fitting}, where only the 
total rank budget and the matrix to be approximated are given.
We present variations and extensions to MLR in \S\ref{s-extension}.
Finally we illustrate our methods with 
numerical examples in \S\ref{s-examples}.

\section{Multilevel low rank matrices} \label{s-MLR}

In this section we define multilevel low rank (MLR) matrices and 
set our notation. 
We also give a few basic properties of MLR matrices, even though our focus is on
the problem of fitting or approximating a matrix with an MLR one.

\subsection{Contiguous multilevel low rank matrices}\label{s-contig-MLR}
We first describe a special case of MLR matrices, one where
the index partitions are contiguous, which simplifies the notation.
An $m \times n$ contiguous MLR matrix $A$ with $L$ levels has the form
\BEQ\label{e-mlr-a}
A = A^1 + \cdots + A^L,
\EEQ
where 
\[
A^l = \blkdiag(A_{l,1}, \ldots, A_{l,p_l} ), \quad l=1, \ldots , L,
\]
where $\blkdiag$ (or direct sum) 
concatenates its not necessarily square matrix arguments 
in both rows and columns.  
Here $p_l$ is the size of the partition at level $l$, with $p_1=1$.
We refer to $A_{l,k}$ as the $k$th block on level $l$.
The dimensions of the block $A_{l,k}$ are $m_{l,k} \times n_{l,k}$, for
$l=1, \ldots, L$, $k=1, \ldots, p_l$.  
For the sum above to make sense, we must have
\[
\sum_{k=1}^{p_l} m_{l,k} = m, \quad
\sum_{k=1}^{p_l} n_{l,k} = n, \quad
l=1,\ldots, L.
\]

The block dimensions on level $l$ partition the row and column indices
into $p_l$ groups, which are contiguous. Since $p_1=1$, the level $1$ row 
and column partitions each consist of one set,
\[
I = \{1, \ldots, m\}, \qquad J = \{1, \ldots, n\}.
\]
Less trivially, the level $2$ partition of the row indices is the set of $p_2$ 
index sets
\[
\{1, \ldots, m_{2,1}\}, ~
\{m_{2,1}+1, \ldots, m_{2,1}+ m_{2,2}\}, ~\ldots,~
\{m-m_{2,p_2}+1, \ldots, m\}.
\]
We require that these partitions be hierarchical, which means that the 
row or column partition at level $l$ is a refinement of the row or column 
partition at level $l-1$, for $l=2, \ldots, L$.

We require that blocks on level $l$ have rank not exceeding $r_l$,
for $l=1, \ldots, L$.  We write them in factored form as 
\[
A_{l,k} = B_{l,k} C_{l,k}^T, \quad B_{l,k} \in \reals^{m_{l,k}\times r_l}, \quad
C_{l,k} \in \reals^{n_{l,k}\times r_l}, \quad
l=1, \ldots, L, \quad k=1, \ldots, p_l,
\]
and refer to $B_{l,k}$ and $C_{l,k}$ as the left and right factors (of block $k$ on level $l$).
We refer to $r= r_1+ \cdots + r_L$ as the MLR-rank of $A$.
The MLR-rank of $A$ is in general not the same as the rank of $A$.
We refer to $(r_1, \ldots, r_L)$ as the rank allocation, \ie, how the total rank $r$
is divided across the levels.
The factors $B_{l,k}$ and $C_{l,k}$ are of course not unique; we can replace them
with $B_{l,k}E$ and $C_{l,k}E^{-T}$ for any invertible $r_l\times r_l$ matrix $E$,
and we obtain the same matrix $A$.



\paragraph{Example.}
To illustrate our notation we give an example with $L=3$ levels, 
with the second level partitioned into $p_2=2$ groups, and the third level
partitioned into $p_3=4$ groups.
We take $m=10$ and $n=8$, with block row dimensions 
\[
\begin{array}{l}
m_{1,1}=10\\
m_{2,1}=4, \quad
m_{2,2}=6,  \\
m_{3,1}=2, \quad
m_{3,2}=2, \quad
m_{3,3}=4, \quad
m_{3,4}=2,
\end{array}
\]
and block column dimensions
\[
\begin{array}{l}
n_{1,1}=8\\
n_{2,1}=4, \quad
n_{2,2}=4,  \\
n_{3,1}=2, \quad
n_{3,2}=2, \quad
n_{3,3}=2, \quad
n_{3,4}=2.
\end{array}
\]
The sparsity patterns of $A^2$ and $A^3$ are shown below,
with $*$ denoting a possibly nonzero entry, and all other entries zero.
(The sparsity pattern of $A^1$ is full, \ie, all entries are possibly nonzero.)
The colors indicate how the partition is being refined when going from level 
$l=2$ to level $l=3$.
\[
A^2 = \left[ \begin{array}{cccccccc}
{\color{magenta}*}&{\color{magenta}*}&{\color{magenta}*}&{\color{magenta}*} \\
{\color{magenta}*}&{\color{magenta}*}&{\color{magenta}*}&{\color{magenta}*} \\
{\color{magenta}*}&{\color{magenta}*}&{\color{magenta}*}&{\color{magenta}*} \\
{\color{magenta}*}&{\color{magenta}*}&{\color{magenta}*}&{\color{magenta}*} \\
&&&&{\color{cyan}*}&{\color{cyan}*}&{\color{cyan}*}&{\color{cyan}*}\\
&&&&{\color{cyan}*}&{\color{cyan}*}&{\color{cyan}*}&{\color{cyan}*}\\
&&&&{\color{cyan}*}&{\color{cyan}*}&{\color{cyan}*}&{\color{cyan}*}\\
&&&&{\color{cyan}*}&{\color{cyan}*}&{\color{cyan}*}&{\color{cyan}*}\\
&&&&{\color{cyan}*}&{\color{cyan}*}&{\color{cyan}*}&{\color{cyan}*}\\
&&&&{\color{cyan}*}&{\color{cyan}*}&{\color{cyan}*}&{\color{cyan}*}
\end{array}\right], \qquad
A^3 = \left[ \begin{array}{cccccccc}
{\color{magenta}*}&{\color{magenta}*} \\
{\color{magenta}*}&{\color{magenta}*} \\
&& {\color{magenta}*}&{\color{magenta}*} \\
&& {\color{magenta}*}&{\color{magenta}*} \\
&&&& {\color{cyan}*}&{\color{cyan}*} \\
&&&& {\color{cyan}*}&{\color{cyan}*} \\
&&&& {\color{cyan}*}&{\color{cyan}*} \\
&&&& {\color{cyan}*}&{\color{cyan}*} \\
&&&&&& {\color{cyan}*}&{\color{cyan}*} \\
&&&&&& {\color{cyan}*}&{\color{cyan}*} 
\end{array}\right].
\]
If we have ranks $r_1=2$, $r_2=1$, and $r_3=1$, the MLR-rank is $r=4$.
This means that $A^1$ has rank $2$, the $p_2=2$ blocks in $A^2$ each have rank $1$,
and the $p_3=4$ blocks in $A^3$ also have rank $1$.
(Of course in practice we are interested in much larger matrices.)

\subsection{Two-matrix forms}
A contiguous MLR matrix is specified by its left and right factors $B_{l,k}$ and $C_{l,k}$,
for $l=1, \ldots, L$ and $k=1, \ldots, p_l$.
These factors can be arranged into two matrices in several ways.

\paragraph{Factor form.}
For each level $l=1, \ldots, L$ define 
\[
\tilde B_l = \blkdiag(B_{l,1}, \ldots, B_{l,p_l}) \in \reals^{m \times p_lr_l}, \qquad
\tilde C_l = \blkdiag(C_{l,1}, \ldots, C_{l,p_l}) \in \reals^{n \times p_lr_l}.
\]
Then we have 
\[
A^l = \tilde B_l \tilde C_l^T, \quad l=1, \ldots, L.
\]
Define 
\[
\tilde B = \left[ \begin{array}{ccc} 
\tilde B_1 & \cdots & \tilde B_L
\end{array}\right] \in \reals^{m \times s}, \qquad
\tilde C = \left[ \begin{array}{ccc} 
\tilde C_1 & \cdots & \tilde C_L
\end{array}\right]\in \reals^{n \times s},
\]
with $s = \sum_l p_l r_l$.
Then we can write $A$ as
\[
A = \tilde B \tilde C^T,
\]
exactly the form of a low rank factorization of $A$.
But here $\tilde B$ and $\tilde C$ have $s$ columns, and a very specific 
sparsity structure, with column blocks that are block diagonal.

\paragraph{Compressed factor form.}
We can also arrange the factors into two arrays or dense matrices with 
dimensions $m \times r$ and $n \times r$.
We vertically stack the factors at each level to form matrices
\[
B^l = \left[ \begin{array}{c} 
B_{l,1} \\ \vdots \\ B_{l,p_l}
\end{array} \right] \in \reals^{m \times r_l}, \qquad
C^l = \left[ \begin{array}{c} 
C_{l,1} \\ \vdots \\ C_{l,p_l}
\end{array} \right] \in \reals^{n \times r_l}, \quad l=1, \ldots, L.
\]
We horizontally stack these matrices to obtain two matrices
\[
B = \left[ \begin{array}{ccc}
B^1 & \cdots & B^{L}
\end{array} \right] \in \reals^{m \times r}, \qquad
C = \left[ \begin{array}{ccc}
C^1 & \cdots & C^{L}
\end{array} \right] \in \reals^{n \times r}.
\]
All of the coefficients in the factors of a contiguous MLR matrix are contained 
in these two matrices.  To fully specify a contiguous MLR matrix, we need to give
the block dimensions $m_{l,k}$ and $n_{l,k}$ for $l=1, \ldots, L$, $k=1, \ldots, p_l$,
and the ranks $r_1, \ldots, r_L$.
We refer to the matrices $B$ and $C$, together with these dimensions, as the two-matrix
form of a contiguous MLR matrix.

For future use, we observe that the matrix $A$ is a bi-linear function of $B$ and $C$,
\ie, it is a linear function of $B$ for fixed $C$, and vice versa.

\subsection{Multilevel low rank matrices}
A general $m \times n$ MLR matrix $\tilde A$ has the form
\[
\tilde A = P A Q^T,
\]
where $A$ is a contiguous MLR matrix, 
$P \in \reals^{m \times m}$ is the row permutation matrix, and
$Q \in \reals^{n \times n}$ is the column row permutation matrix.
To specify an MLR matrix $\tilde A$ we specify the two matrices $B\in \reals^{m \times r}$
and $C\in \reals^{n\times r}$,
the block dimensions, the ranks, and finally, the permutations $P$ and $Q$.

We can think of an MLR matrix as using a general hierarchical partition of the
row and column index sets $I$ and $J$, with the level $l$ partition containing
$p_l$ groups.
In contrast, a contiguous MLR matrix uses hierarchical row and column partitions
where the groups in the partitions are contiguous ranges.

\paragraph{Low rank matrix.}
When $L=1$, an MLR matrix $A$ is simply a matrix with rank no more than $r$,
say $A=BC^T$ with $B\in \reals^{m \times r}$ and $C\in \reals^{n \times r}$.
In this case matrix rank and MLR matrix rank agree.
The associated two-matrix form uses $B$ and $C$.
So MLR matrices generalize low rank matrices.
We will see that low rank matrices of rank $r$ and MLR matrices of rank $r$
share several attributes.  For example both require the storage of 
$(m+n)r$ real coefficients.

\subsection{Variations}
We will consider two variations on MLR matrices.

\paragraph{Symmetric MLR matrices.}
Here we consider the special case where the matrix $A$ is symmetric.
In a symmetric MLR matrix, we require that the row and column permutations are 
the same, \ie, $P=Q$,
and all blocks $A_{l,k}$ are symmetric.  This implies that
$m_{l,k}=n_{l,k}$ for all $l$ and $k$.
The left and right factors can be chosen to be the same,
up to a sign.  That is we take $C_{l,k} = B_{l,k} S_{l,k}$, where $S_{l,k}$ is a 
diagonal sign matrix. This means we only store the matrix $B$ (say), together with the
signs in $S_{l,k}$.
For a symmetric MLR matrix, we need to store only half the number of coefficients
as a general square MLR matrix of the same size and rank.

\paragraph{Positive semidefinite MLR matrices.}  
Here we consider a further restriction beyond symmetry:
$A$, and each block $A_{l,k}$, is symmetric positive semidefinite (PSD).
In this case we can take $B_{l,k}=C_{l,k}$.

\paragraph{Example: Factor covariance matrix.}
A common model for an $n \times n$ covariance matrix is the so-called factor model,
\[
\Sigma = F F^T + D,
\]
where $F\in \reals^{n \times p}$ is the factor loading 
matrix and $D$ is a diagonal PSD matrix.  The dimension $p$ is referred to 
as the number of factors in the model.
This common model is exactly a PSD MLR matrix, with $L=2$,
and $n_{2,k}=1$ for $k=1, \ldots, n$, $r_1 = p$ and $r_2=1$,
so as an MLR matrix, $\Sigma $ has rank $r=p+1$.
We take $B_{1,1}=F$, and $B_{2,k} = \sqrt{D_{kk}}$, $k=1, \ldots, n$.
(With this specific hierarchical partitioning, the permutation $P$ does not matter,
since for any permutation we obtain the same form.)

\subsection{Properties}
Here we briefly describe a few of the nice properties of MLR matrices.  
But this is not our focus,
which is on fitting MLR matrices, considered in the next section.

\paragraph{Matrix-vector multiply.}
Suppose $A$ is MLR with rank $r$, and we wish to evaluate $Ax$.
The pre- and post-permutations 
require no floating point operations, so we can just assume that $A$ is contiguous
MLR.  To evaluate $Ax$, we evaluate $A_{l,k}x_{l,k}$ for $l=1,\ldots,L$, $k=1, \ldots, p_l$,
where $x_{l,k}$ is the appropriate subvector or slice of $x$.  Each of these
matrix-vector multiplies are carried out in an obvious way by first evaluating
$z_{l,k}=C_{l,k}^Tx_{l,k}$, and then forming $A^{l,k}x_{l,k} = B_{l,k}z_{l,k}$.
These require around 
$2n_{l,k}r_l$ and $2m_{l,k}r_l$ flops each, so the total is around
\[
\sum_{l=1}^L \sum_{k=1}^{p_l} 2(n_{l,k}+m_{l,k})r_l =
\sum_{l=1}^L 2(n+m)r_l = 2(n+m)r
\]
flops in total. (In addition there is much opportunity for carrying out the
block matrix-vector multiplies in parallel.)
This is the same flop count for evaluating $Ax$ when $A=BC^T$ is a rank $r$ matrix,
again illustrating the idea that MLR matrices are generalizations of low rank
matrices.

Evidently we can carry out the adjoint multiply $A^Ty$ with the same complexity.
Indeed, if $A$ is MLR, so is $A^T$.  To get the MLR representation of $A^T$ from that
$A$, we swap $P$ and $Q$, $B$ and $C$, and the 
dimensions $m_{l,k}$ and $n_{l,k}$. The ranks $r_l$ are the same, so the
rank of $A^T$ (in the MLR sense) is the same as of $A$.

\paragraph{Solving equations and least squares.}
Solving systems of linear equations, least squares problems, and equality 
constrained least squares problems that involve MLR matrices can be done
using methods that require only matrix-vector and matrix-transpose-vector
multiplies, exploiting the efficiency of those operations.
Alternatively, they can be reformulated as solving systems of sparse linear
equations.

Consider the system of linear equations $Ax=b$, with $A$ invertible and MLR.
Using the factor form $A=\tilde B \tilde C^T$,
and introducing a new variable $z \in \reals^s$, 
we get two new systems of equations
\BEQ\label{e-exp1}
\tilde B z = b, \qquad \tilde C^T x = z.
\EEQ
In replacing the system of equations $Ax=b$ with \eqref{e-exp1},
we have introduced a total of $s=\sum_{l=1}^L r_l p_l$
new scalar variables.  
But each system of equations in \eqref{e-exp1} is sparse, 
and its sparsity pattern can be exploited by a sparse direct solver~\cite{davis2016survey}.

The solution of the least squares problem of choosing $x$ to minimize $\|Ax-b\|_2^2$,
is given by solving a system of normal equations $A^T(Ax - b)=0$. 
Now consider the case where $A$ is MLR.
We introduce new variables $z_1, z_3 \in \reals^s$, $z_2 \in \reals^m$, 
and get an equivalent new system of equations
\BEQ\label{e-exp2}
\begin{bmatrix}
\tilde C^T & -I & 0 & 0 \\
0 & \tilde B & -I & 0 \\
0 & 0 & \tilde B^T & -I \\
0 & 0 & 0 & \tilde C
\end{bmatrix} 
\begin{bmatrix}
x \\
z_1 \\
z_2 \\
z_3 \\
\end{bmatrix} = 
\begin{bmatrix}
0 \\
0 \\
0 \\
\tilde C \tilde B^Tb \\
\end{bmatrix}.
\EEQ 

Note that the system of equations~\eqref{e-exp2} is sparse  
and thus a sparse direct solver, as previously mentioned~\cite{davis2016survey}, 
can be employed.
This process of expanding a system to a large sparse system is called 
extended sparsification~\cite{chandrasekaran2007fast, 
pouransari2017fast}. 

Lastly, consider the constrained least squares problem
\BEQ\label{e-constr-ls-prob}
\begin{array}{ll}
\mbox{minimize} & \|Ax - b\|_2^2 \\
\mbox{subject to} & Gx = h,
\end{array}
\EEQ
where $A$ and $G$ are MLR matrices. The optimality (KKT) conditions 
for~\eqref{e-constr-ls-prob} are
\BEQ\label{e-exp3}
\begin{bmatrix}
2A^TA & G^T \\
G & 0
\end{bmatrix} 
\begin{bmatrix}
x \\
y
\end{bmatrix} = 
\begin{bmatrix}
2A^T b\\
h
\end{bmatrix}.
\EEQ 
Given that both $A$ and $G$ are MLR, the system~\eqref{e-exp3} 
can be sparsified as in~\eqref{e-exp2} and solved efficiently afterwards.

\section{MLR fitting problem}\label{s-fitting}
\subsection{General MLR fitting}
The most general MLR fitting problem is
\BEQ\label{e-fitting-prob}
\begin{array}{ll}
\mbox{minimize} & \|A - \hat A\|_F^2 \\
\mbox{subject to} & \hat A~\mbox{is rank $r$ MLR},
\end{array}
\EEQ
where $A \in \reals^{m \times n}$ is the given matrix to be fit,
$\hat A$ is the variable, and $\| \cdot \|_F^2$ denotes the square of the 
Frobenius norm, \ie, the sum of squares of the entries.
In this general version of the problem the data are $A$, the matrix to be fit,
and $r$, the rank.  We seek permutations $P$ and $Q$, 
the number of levels $L$, the block dimensions 
$m_{l,k}$ and $n_{l,k}$, $l=1, \ldots, L$, $k=1, \ldots, p_l$, the
two matrices $B$ and $C$, 
and the rank allocation, \ie,
$r_i$ for which $r_1 + \cdots + r_L=r$.

\paragraph{Variations.}
We consider several variations on the general 
MLR fitting problem \eqref{e-fitting-prob}.
In the symmetric MLR fitting problem, we assume that $m=n$, the matrix $A$
is symmetric, and we require that $\hat A$ be a symmetric MLR matrix.
In the PSD MLR fitting problem, we assume in addition that $\hat A$ is a 
PSD MLR matrix, and require that $\hat A$ is a PSD MLR matrix.

\subsection{Factor fitting and rank allocation}
We also consider variations in which some parameters in the general MLR fitting
problem \eqref{e-fitting-prob} are fixed, which means we 
optimize over fewer parameters.
Each of these variations can also be restricted to symmetric or PSD MLR matrices.

\paragraph{Factor fitting problem.}
In the factor fitting problem
we fix the hierarchical partition, \ie, the permutations 
$P$ and $Q$, the number of levels $L$, and the block dimensions $m_{l,k}$ and $n_{l,k}$,
and we also fix the rank allocation, \ie, $r_1, \ldots, r_L$.
We optimize over the factors, \ie, the matrices $B$ and $C$.
Roughly speaking, in the factor fitting problem we fix the combinatorial aspects of 
$\hat A$, and only optimize over the (real) coefficients in the factors.

\paragraph{Rank allocation problem.}
In the rank allocation problem we fix the hierarchical partition but not the ranks.
This means we optimize over $B$ and $C$, and also the ranks $r_1,\ldots, r_L$, subject to
$r_1+\cdots+r_L=r$.
The rank allocation problem is a natural one when the hierarchical row and column
partitions are given or already known.

\paragraph{Complexity.}
Aside from the special case of low rank matrices described below,
we do not hope to solve the fitting problem exactly,
but only develop good heuristic methods for it.
In the next section we describe a method for the most 
constrained problem, factor fitting. In the two following sections we extend 
that to rank allocation, and
the full fitting problem \eqref{e-fitting-prob}.

\subsection{Special case: Low rank matrix}
We can solve the MLR fitting problem exactly when there is one level, so rank $r$ MLR 
reduces to rank $r$.  The well-known solution is found from the
singular value decomposition (SVD) of $A$, $A=U\Sigma V^T$.
An optimal rank $r$ approximation is $\hat A = BC^T$,
\[
B = U_r \Sigma_r^{1/2}, \qquad
C^T = \Sigma_r^{1/2} V_r,
\]
where $U_r$ is the first $r$ columns of $U$,
$V_r$ is the first $r$ columns of $V$,
and $\Sigma_r$ is the leading $r\times r$ submatrix of $\Sigma$.
The optimal (minimum) objective value is $\sum_{i=r+1}^{\min\{m,n\}} \sigma_i^2$.

The symmetric and PSD versions of the low rank matrix fitting problem
can also be solved exactly.  Here we use 
an eigendecomposition of $A$, $A=Q\Lambda Q^T$, with $Q$ an orthogonal matrix
of eigenvectors and $\Lambda$ a diagonal matrix of eigenvalues, sorted from largest
to smallest in absolute value.  In the symmetric case we take $B=Q_r|\Lambda_r|^{1/2}$
and $C=BS$, where $Q_r$ is the leading $n \times r$ submatrix of $Q$, 
$\Lambda_r$ is the leading $r \times r$ submatrix of $\Lambda$, and $S$ is a 
diagonal $r \times r$ sign matrix, with $S_{ii} = \sign \Lambda_{ii}$.
In the PSD case, we use the eigendecomposition, with the entries of $\Lambda$ sorted
from largest to smallest.  We replace $\Lambda_r$ in the symmetric case with
$\max(\Lambda_r,0)$, elementwise.
We will leverage these analytical solutions for the special case of low rank
matrices in the methods we propose for solving the general MLR fitting problem.

\paragraph{Partial SVDs.}
The singular value and eigenvalue decompositions described above can be computed 
in full, which has a complexity of order $\min\{mn^2,m^2n\}$ flops. 
In most cases, however, the target rank is substantially smaller than
$\min\{m,n\}$, in which case methods for computing only a set of extremal
singular values or eigenvalues and their associated singular vectors and 
eigenvectors can be used \cite{sorensen1992implicit, golub1996, lehoucq1998arpack}.  
These methods have complexity 
that can be far smaller than computing a full SVD or eigendecomposition.
(They can also be warm-started, which is an advantage in the 
context of the methods we describe later.)

\paragraph{Alternating least squares.}
We mention for future use that the best rank $r$ approximation can
also be computed using alternating least squares, applied to the 
objective
\[
\|A-BC^T\|_F^2,
\]
with variables $B\in \reals^{m \times r}$ and $C\in \reals^{n \times r}$.
The method alternates between minimizing the objective over $B$, with $C$ fixed,
and then minimizing the objective over $C$, with $B$ fixed.  
These are both simple least squares problems with analytical formulas for the solution.
Moreover when we minimize over $C$, the least squares problem splits into 
separate ones for each row of $C$, which can be solved in parallel, and similarly
for $B$.

\clearpage
\section{Factor fitting methods}\label{s-fitting-methods}
In the factor fitting problem we are given $A$, the matrix 
to be approximated, and for the MLR approximator $\hat A$, we are 
given the permutations $P$ and $Q$, the number of levels $L$,
the block dimensions, and the ranks.  We are to choose the 
matrices $B$ and $C$, \ie, the factors $B_{l,k}$ and $C_{l,k}$ to
minimize $\|A-\hat A \|_F^2$.
Since $P$ and $Q$ are given, we can just as well fit 
$\tilde A = P^TAQ$ with a contiguous MLR matrix $\hat A$.
We can also have an initial guess of the factors.

\subsection{Alternating least squares}\label{s-als}



Let $B$ and $C$ be the two matrices associated with the MLR-rank $r$ approximation
$\hat A$.  We write this as $\hat A(B,C)$, and recall that $\hat A$ is bi-linear, 
\ie, using the factor form we have $\hat A = \tilde B \tilde C^T$.
We can use an alternating least squares (ALS) algorithm to minimize 
\BEQ\label{e-als-obj}
\|A - \hat A(B,C)\|_F^2
\EEQ
over $B$, then $C$, then $B$, etc. 

There are many possible ways to carry out the minimization over $C$ (with fixed $B$).
We can get a closed-form 
solution for the least squares problems for each row of $\tilde C$
while taking into account the sparsity of $\tilde C$.
Another approach is to use an iterative method such as conjugate gradients (CG) 
applied to the normal equations associated with minimizing~\eqref{e-als-obj}.
An iterative approach allows us to solve the minimization over $B$ or $C$ 
approximately, which is appropriate since the least squares problems
solved are just one step in an iterative algorithm.

\begin{algdesc}{\sc Alternating least squares for factor fitting}\label{alg-als-ff}
    {\footnotesize
    \begin{tabbing}
    {\bf given} $A \in \reals^{m \times n}$, permutations $P$, $Q$,
    block sizes $m_{l,k}$ and $n_{l,k}$, ranks
    $r_1, \ldots,r_L$,\\
and initial factors $B_{l,k}$ and $C_{l,k}$, $l=1, \ldots, L$,
$k=1, \ldots, p_l$.  \\*[\smallskipamount]
    \emph{Permute to contiguous form.} Form $\tilde A =P^TAQ$.
\\*[\smallskipamount]
    {\bf for iteration/epoch $t=1,2,\ldots$} \\*[\smallskipamount]
    \emph{Check stopping criterion.} Quit if~\eqref{e-stopping-crit}
holds.\\
     Approximately minimize \eqref{e-als-obj} over $B$. \\
     Approximately minimize \eqref{e-als-obj} over $C$. 
    \end{tabbing}}
\end{algdesc}

We have experimented with several methods for approximate minimization 
of~\eqref{e-als-obj} and found that taking $10$ steps of CG performs well
over a wide range of problems.
This alternative least squares method converges to $B$ and $C$ that 
are locally optimal, in the sense that no other choice of $B$ will
decrease the objective for the given $C$, and vice versa.
But it need not be a global solution to the factor fitting problem.

\subsection{Block coordinate descent}\label{s-bcd}
We can use a block coordinate descent (BCD) algorithm, updating the
factors in one level in each iteration.  As we will see, we can 
carry out exact minimization over the factors on any one level.
To optimize over the factors in level $l$, we need to choose $B_{l,k}$ and $C_{l,k}$
to minimize
\BEQ\label{e-blk-obj}
\left\| R - 
 \blkdiag(B_{l,1}C_{l,1}^T, \ldots, B_{l,p_l}C_{l,p_l}^T) \right\|_F^2,
\EEQ
where 
\[
R = \tilde A - \sum_{j \neq l} 
\blkdiag(B_{j,1}C_{j,1}^T, \ldots, B_{j,p_j}C_{j,p_j}^T)
\]
is the current residual.
This problem separates into $p_l$ independent problems, where we minimize
$\|R_{l,k} - B_{l,k}C_{l,k}^T \|_F^2$, where $R_{l,k}$ is the submatrix of $R$
corresponding to the $l,k$ block.
But this we know how to do exactly, as described above, using the SVD of $R_{l,k}$,
or the eigenvalue decomposition for the symmetric and PSD cases.
This step can be carried out in parallel for all $p_l$ blocks in level $l$.
Moreover we can use SVD methods that allow us to compute only an appropriate
number of the largest singular values and vectors, and not a full SVD \cite{golub1996}.

There are many possible choices for the order in which we update the levels.
We have found that an effective method is to start from level~$l=1$, proceed 
in order to level $L$, and then in reverse order update levels from $l=L-1$ back to 
$l=1$.  In many cases one such V-shape sweep, which we refer to as an epoch, 
suffices; if not, it can be repeated.
We stop when the relative fit 
does not improve much over an epoch, \ie,
\BEQ\label{e-stopping-crit}
\frac{\|A-\hat A^\mathrm{prev}\|_F}{\|A\|_F} -
\frac{\|A-\hat A\|_F}{\|A\|_F}\leq \epsilon_{\text{rel}}\frac{\|A-\hat A^\mathrm{prev}\|_F}{\|A\|_F},
\EEQ
where $\epsilon_{\text{rel}}$ is a positive parameter, $\hat A$ is the current approximation,
and $\hat A^\mathrm{prev}$ is the approximation 
in the previous epoch.

\begin{algdesc}{\sc Block coordinate descent for factor fitting}\label{alg-bcd-ff}
    {\footnotesize
    \begin{tabbing}
    {\bf given} $A \in \reals^{m \times n}$, permutations $P$, $Q$,
    block sizes $m_{l,k}$ and $n_{l,k}$, ranks
    $r_1, \ldots,r_L$,\\
and initial factors $B_{l,k}$ and $C_{l,k}$, $l=1, \ldots, L$,
$k=1, \ldots, p_l$.  \\*[\smallskipamount]
    \emph{Permute to contiguous form.} Form $\tilde A =P^TAQ$.
\\*[\smallskipamount]
    {\bf for iteration/epoch $t=1,2,\ldots$} \\*[\smallskipamount]
    \emph{Check stopping criterion.} Quit if~\eqref{e-stopping-crit}
holds.\\
     \emph{Carry out V-epoch.} {\bf for $l=1,\ldots,L-1,  L, L-1, \ldots, 1$} \\
     \qquad \emph{Form current residual.} $R=\tilde A -
\sum_{j\neq l} \blkdiag(B_{j,1}C_{j,1}^T, \ldots, B_{j,p_j}C_{j,p_j}^T)$.\\
     \qquad Update $B_{l,k}$ and $C_{l,k}$ for $k=1, \ldots, p_l$ by minimizing
\eqref{e-blk-obj}.
    \end{tabbing}}
\end{algdesc}

This is a descent method and converges to at least a local minimum.
We have observed that, depending on the initialization,
it can converge to different local minima, even with somewhat different 
final objective values.   We have also observed that the BCD method
works well when all factors are initialized at $0$, when no better initial 
guess is available.


\paragraph{Example: Factor covariance fitting.} The BCD algorithm takes an interesting form
when applied to the factor covariance fitting problem.  Here $A$ is the empirical 
covariance matrix and $\hat A$ will have the form $\hat A = FF^T + \diag(d)$, with $F\in 
\reals^{n \times p}$ and $d$ nonnegative.
If we initialize with $F=0$ and $d=0$ the first four BCD steps are
\begin{enumerate}
	\item Set $F$ as the rank-$p$ PSD approximant of $A$.
	\item Set $d = \diag(A)-\diag(FF^T)$.
	\item Set $F$ as the rank-$p$ PSD approximant of $A-\diag(d)$.
	\item Set $d = \max(\diag(A)-\diag(FF^T),0)$.
\end{enumerate}
In step~2 we are guaranteed that $d$ is nonnegative, since $A-FF^T$ is PSD.
Steps~1 and~2 are the typical method for fitting a factor model to a given
covariance matrix $A$.  The choice of $d$ guarantees that $A$ and $\hat A$ agree on
the diagonal entries, \ie, $\hat A$ captures exactly the (marginal) variance of 
each component.
Steps~3 and~4 (and more, if the iteration continues) are not traditional,
but can further reduce the Frobenius approximation error.

\subsection{Comparison}

In this section we show some typical convergence results for the 
ALS and BCD fitting methods, using as an example a matrix $A$
of size $m=5000$ and $n=7000$, fitting with an MLR matrix with $L=14$ levels
and ranks $r_1=\cdots=r_{14}=5$.
(The matrix is a discrete Gauss transform, described in \S\ref{s-dgt}.)
The matrices $B$ and $C$ are initialized 
as the matrices resulting from a single sweep of BCD from $l=1$ to $l=14$.

\begin{figure}
    \begin{center}    
    \includegraphics[width=0.8\textwidth]{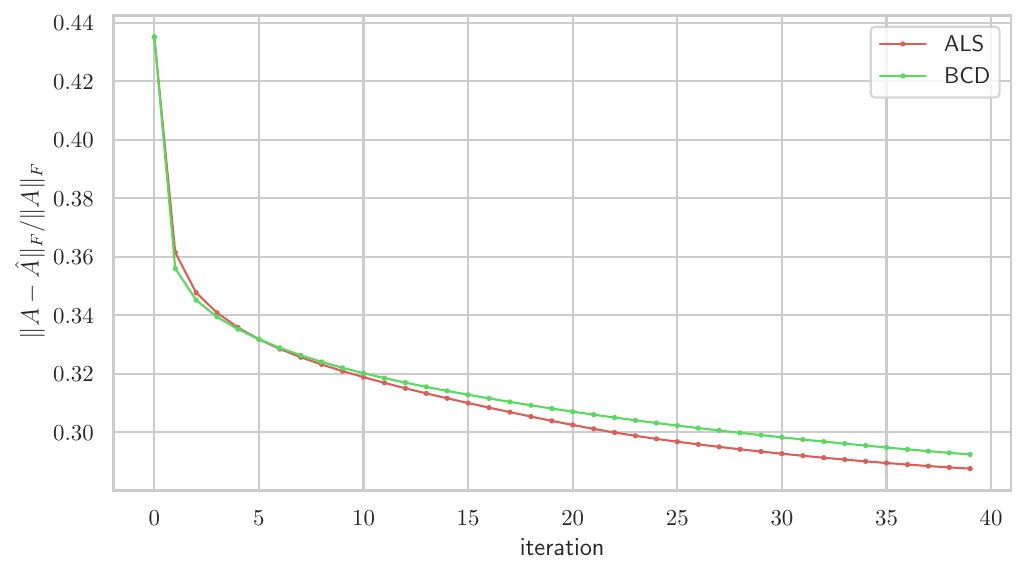}
    \end{center}
    \caption{Comparison of BCD and ALS. For ALS, one iteration is approximately
minimizing over $B$ and then over $C$. For BCD, one iteration is one V-epoch.}
    \label{fig-bcd-als-comp}
\end{figure}

Figure~\ref{fig-bcd-als-comp} shows the relative fitting loss versus
iterations. 
For the ALS method, one iteration is approximately minimizing over $B$
and then over $C$ (using 10 steps of CG).
For the BCD method, one iteration is one V-epoch.
The convergence of the two methods is quite similar, at least in terms
of iterations.  

The total time, however, is very much dependent on 
the implementation, how much parallelism is exploited, and the 
computing platform used.
Our non-optimized implementation uses SciPy~\cite{2020SciPy-NMeth} for the
partial SVD in BCD, and
PyTorch~\cite{paszke2019pytorch} to compute the gradient of the quadratic 
loss function~\eqref{e-als-obj} during the CG step of ALS.
It requires on the order of $10$s of seconds for each iteration of ALS or BCD.


\clearpage
\section{Rank allocation}\label{s-rank-alloc}
In this section we describe a method for adjusting the ranks, subject to 
$r_1+ \cdots + r_L=r$, in order to reduce the Frobenius norm 
of the approximation error.

\subsection{Incrementing and decrementing level rank}
We start with any valid allocation of rank, such as $r_1=r$ and $r_2 = \cdots = r_L=0$.
We use the factor BCD method above to optimize the factors for this rank allocation.
(Results with ALS are similar.)
We then find an approximation of how much better the approximation would be,
if we were to increase the rank allocated to level $l$ by one.  We also
obtain an approximation of how much worse the approximation would be if we were to
decrease the rank allocated to level $l$ by one.  (If $r_l=0$ it is 
infeasible to reduce the rank allocated to it.)

To do this, recall that the minimum Frobenius norm squared approximation of a matrix
by a rank $r$ one is $\sum_{i=r+1}^{\min\{m,n\}} \sigma_i^2$, where $\sigma_i$ are 
the singular values of the matrix.
If we increase the rank of the approximation by one (assuming $r<\min\{m,n\}$),
to $r+1$, it follows that
the Frobenius norm square fitting error decreases by $\sigma_{r+1}^2$.
If we decrease the rank of the approximation by one (assuming $r>0$) 
to $r-1$, it follows that the fitting error increases by $\sigma_{r}^2$. 
If we increase the rank allocated to (each block of)
level $l$ by $1$, the decrease in Frobenius norm squared error is
\[
\delta_l^+ = \sum_{k=1}^{p_l} \sigma_{r_l+1}^2 (R_{l,k}).
\]
If we reduce the rank allocated to level $l$, the increase in Frobenius norm
squared error is
\[
\delta_l^- = \sum_{k=1}^{p_l} \sigma_{r_l}^2 (R_{l,k}).
\]
These numbers predict the decrease or increase when only the level $l$ factors
are changed; by using BCD, the Frobenius norm squared error can only decrease.
Nevertheless we have found these numbers to give good enough estimates of the changes
for our rank allocation algorithm to be effective.

\subsection{Rank exchange algorithm}
In each iteration of the rank allocation algorithm, we find the level $i$ and $j$, with
$i\neq j$, for which the predicted net decrease is maximized, \ie, 
\[
	i,j = \argmax_{i \neq j} \left( \delta^+_i - \delta^-_j\right).
\]
We then carry out this suggested rank re-allocation or rank exchange, by 
increasing $r_i$ by one and decreasing $r_j$ by one.  
We then use the BCD factor fitting method, in warm start mode,
to update the factors to further decrease the fitting error.  
We reject the update if the fitting error
with the new rank allocation is worse than the current one, and quit.

The PSD version of this method tends to converge faster,
since it often occurs that $\delta_l^+=0$, \ie, allocating more rank to
level $l$ would not improve the current approximation error.
(This occurs when the current level has a lower rank than is allocated to it.)

\begin{figure}
    \begin{center}    
    \includegraphics[width=0.8\textwidth]{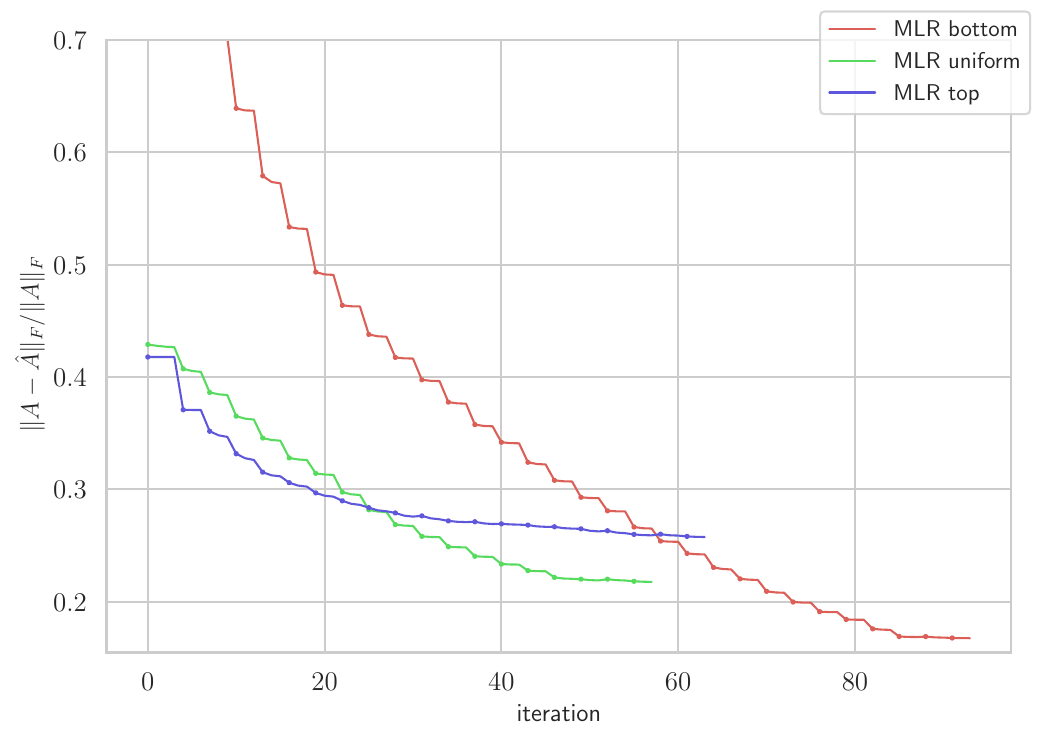}
    \end{center}
    \caption{Fitting error during rank allocation, starting from three different
initial allocations of rank.}
    \label{fig-loss-dgt}
\end{figure}

\paragraph{Example.}
Figure~\ref{fig-loss-dgt} shows some typical fitting error trajectories 
during rank allocation, for a matrix with $m=5000$ and $n=7000$,
where we allocate a total rank $r=28$ to $L=14$ levels.
The horizontal axis shows BCD iterations. 
The point markers on the curves show the iterations 
in which rank was re-allocated.
We show three trajectories, corresponding to different initializations.
The one labeled MLR bottom has initial allocation of rank
entirely to the bottom level, \ie, $r_{14}=28$, $r_1 = \cdots =r_{13}=0$.
The one labeled MLR uniform has initial allocation of rank uses
$r_1= \cdots = r_{14} = 2$.
The one labeled MLR top has initial allocation of rank
$r_1 = 28$, $r_2=\cdots = r_{14}=0$.
(The particular allocations of rank for these can be seen in
figure~\ref{fig-dgt-ra-evolution}.)
We observe that different rank initializations
converge to different local minima, with different loss values.
For this example, the rank allocation obtained starting from the allocation
of rank to the bottom yields the best fitting error.

\clearpage
\section{A method for general MLR fitting}\label{s-MLR-fitting}
In this section we describe a method for the full MLR fitting problem, where
only $A$, the matrix to be approximated, and a rank $r$, are given.
The method constructs the hierarchy one level at a time, starting from the top level $l=1$.
We first describe the general method.
For simplicity we describe the specific case when each block on one level is split
into two nearly equal sized blocks on the next level, \ie, we use nested dissection
on the rows and columns.  This means $p_l=2^{l-1}$, and $L = \lceil \log_2 \min\{m,n\}\rceil+1$.
(The final approximation may have fewer levels than this initial maximum value.)
The row dimensions $m_{l,k}$ are approximately $m 2^{-(l-1)}$, 
and the column dimensions $n_{l,k}$ are approximately $n 2^{-(l-1)}$.
We start with an initial rank allocation that is nearly uniform across levels, \ie,
$r_l \approx r/L$.

We first form a rank $r_1$ approximation of $A$, and subtract it from $A$ to form a residual
matrix $R$. 

The next step is to partition the rows and columns each into two nearly 
equal sized groups $I_{2,1}$ and $I_{2,2}$, and 
$J_{2,1}$ and $J_{2,2}$,
so as to maximize the 
sum of the squares of the residuals within the two groups,
\[
\sum \left\{ R_{ij}^2 \mid (i,j)\in I_{2,1} \times J_{2,1}~\mbox{or}~
(i,j)\in I_{2,2} \times J_{2,2} \right\}.
\]
The sum here includes about half the pairs $(i,j) \in I \times J$.
We do not need to solve this partitioning problem exactly; we describe below a simple 
spectral method, but any method can be used.
We then permute the rows and columns so the partition is contiguous,
\ie, $I_{2,1} = \{1, \ldots, m_{2,1}\}$ and
$I_{2,2} = \{m_{2,1}+1, \ldots, m\}$ and similarly for the columns indices.
We can interpret this step as finding row and column permutations
that give the best block diagonal Frobenius norm approximation of the residual matrix.
Now we use the BCD factor fitting method to update the factors on levels $l=1$ and $l=2$.

We now apply the same method to each of the $p_2=2$ blocks on level $l=2$.
For each of the two blocks we get the current residual, and then partition the rows and columns
within each block to maximize the sum of squares of residuals within the blocks.
Then we permute rows and columns within each block so the partition is contiguous.
We use the BCD factor fitting method to update the factors on levels $l=1$, $l=2$, and $l=3$.

We continue this way, in each step 
partitioning the rows and columns within each block so as to 
maximize the sum of squares of the current residuals within each group of the 
partition, and then permuting the rows and columns so the partitions are 
contiguous.  Finally, we use BCD factor fitting to update all the factors from the top
to the current level.

When we finish with level $L$, we have our first MLR approximation of $A$.
We can now run rank allocation to further improve the fit.

The same method can be used for the symmetric MLR fitting problem,
using the same partitioning of the rows and columns within each block.

\subsection{Symmetric row and column dissection}
Here we describe a simple heuristic for partitioning the rows and columns of a
matrix, each about equally, so as to maximize the sum of squares of a given
residual matrix.

We start with the symmetric case, where the method is just spectral partitioning.
Spectral partitioning goes back to the early
1970s~\cite{hall1970r, donath1973lower, fiedler1973algebraic} 
and was later popularized in the 1990s through seminal contributions
of~\cite{pothen1990partitioning, simon1991partitioning, barnard1994fast}.
We review this here, mostly to explain the generalization to the non-symmetric 
case later.

For simplicity we assume that $n$ is even. Let $S$ be the elementwise square of the
(symmetric) residual matrix $R$.
We represent the partition as a vector $x \in \{-1,1\}^n$, with $x_i = -1$ meaning
$i$ is in the first group, and $x_i=1$ meaning $i$ is in the second.
The two groups have equal size provided $\ones^Tx=0$.
We observe that 
\[
x^TSx = \sum_{i,j} x_ix_j R_{ij}^2 = \sum_{x_i=x_j} R_{ij}^2 -
\sum_{x_i \neq x_j} R_{ij}^2 =
2 \sum_{x_i=x_j} R_{ij}^2 - \|R\|_F^2.
\]
The problem we wish to solve is
\[
\begin{array}{ll} \mbox{maximize} & x^T Sx\\
\mbox{subject to} & x_i \in \{-1,1\}, \quad i=1, \ldots, n, \quad \ones^Tx=0,
\end{array}
\]
with variable $x$.
Changing the diagonal entries of $S$ only adds a constant to 
the objective, and so does not change the solution.  So we change each diagonal entry of $S$
to be the negative of the sum of the other entries in the row.  This results in
the negative of a Laplacian matrix we denote as $L$, with $L\ones =0$.
Substituting $L$ for $S$ in the problem above (and minimizing since we have changed
the objective sign) yields an equivalent problem,
\[
\begin{array}{ll} \mbox{minimize} & x^T Lx\\
\mbox{subject to} & x_i \in \{-1,1\}, \quad i=1, \ldots, n, \quad \ones^Tx=0.
\end{array}
\]

For any $x$ with $x_i \in \{-1,1\}$ we have $\|x\|_2 = \sqrt n$.
Now we relax the problem to 
\[
\begin{array}{ll} \mbox{minimize} & x^T Lx\\
\mbox{subject to} & \|x \|_2^2 =n, \quad \ones^T x = 0,
\end{array}
\]
with variable $x \in \reals^n$.
We will assume that $L_{ij}<0$ for all $i\neq j$ (which we can ensure
by subtracting a small number from each entry), which implies that the 
second smallest eigenvalue of $L$ is positive.
With this assumption, the solution of the problem above is 
$x= \sqrt n v$, where $v$ is the 
eigenvector of the Laplacian associated with its second smallest eigenvalue.
Since $L\ones=0$, we see that $\ones$ is an eigenvector of $L$, so 
we get $\ones^T x= 0$ automatically.

The final step is to project or round this $x$ back to our original constraints
$x_i \in \{-1,1\}$ while maintaining $\ones^Tx = 0$.
This is done by sorting $x$ and taking $x_i=-1$ for the first $n/2$ entries and
$+1$ for the remaining entries.
(The same method works when $n$ is odd; we can arbitrarily choose the group that has 
one more index than the other.)

It is typically followed with a greedy algorithm, where we consider swapping 
a pair, one from each group, and accepting the new partition if the
objective decreases. Finding the change in cost for all such pairs requires
order $n^2$ flops.

\subsection{Non-symmetric row and column dissection}
We now describe a generalization of spectral partitioning to the case 
where we wish to partition both the rows and columns.  
Again for simplicity 
we assume that $m$ and $n$ are even.
We use $u \in \reals^m$ and $v\in \reals^n$ to denote the row and column
partitions, with $u_i, v_i \in \{-1,1\}$.  The constraints $\ones^Tu=\ones^Tv=0$
guarantee that the partitions of rows and columns have equal size.
We observe that
\[
u^T S v = 
\sum_{i,j} u_iv_j R_{ij}^2 = \sum_{u_i=v_j} R_{ij}^2 -
\sum_{u_i\neq v_j} R_{ij}^2 =
2 \sum_{u_i=v_j} R_{ij}^2 - \|R\|_F^2 =
\|R\|_F^2 - 2 \sum_{u_i \neq v_j} R_{ij}^2.
\]
Our goal is to maximize $u^TSv$, subject to 
$u_i,v_i \in \{-1,1\}$, $\ones^Tu = \ones^Tv=0$.
This is closely related to~\cite{dhillon2001co, zha2001bipartite}
which minimizes the normalized sum of edge weights crossing partitions,
while our method does not incorporate normalization.


As with spectral partitioning, we first modify $S$ to $\tilde S$ in a way
that does not change the solution of the problem, but results in $\tilde S \ones =0$
and $\tilde S^T \ones=0$.  This is analogous to changing the diagonal entries of
$S$ in the symmetric case, so that $S\ones =0$.
We then will minimize $u^TMv$, with $M=-\tilde S$.  (The matrix $M$ is analogous to
$L$ in the symmetric case, but it is not a Laplacian matrix.)
For any $a\in \reals^m$ and $b\in \reals^n$ we have
\[
u^T S v = u^T \tilde S v, \qquad  \tilde S = S-a \ones^T - \ones b^T,
\]
for any $u$ and $v$ with $\ones^T u = \ones^T v=0$.
The choice
\[
 a = (1/n)(S\ones - (\ones^T S \ones/2m)\ones), \qquad 
 b = (1/m)(S^T\ones - (\ones^T S \ones/2n) \ones)
\]
yields the desired result, $\tilde S\ones =0$ and $\tilde S^T\ones=0$.

As in spectral partitioning, we form the relaxed problem
\[
\begin{array}{ll} \mbox{minimize} & u^T M v\\
\mbox{subject to} & \|u \|_2^2 =m, \quad \|v\|_2^2=n, \quad \ones^T u = \ones^Tv = 0,
\end{array}
\]
where $M= - \tilde S$.
The solution of this problem is to choose $-u$ and $v$ to be (scaled)
left and right singular vectors
of $M$ associated with its largest (positive) singular value.
Since $M \ones=0$ and $M^T \ones=0$, these vectors are left and right 
singular vectors of $M$ associated with singular value zero.
Thus we automatically have $\ones^T u = \ones^T v =0$.

The final step is to round $u_i$ and $v_i$ to Boolean values.
We do that setting the $m/2$ smallest entries of $u$ to be $-1$, and the remaining ones
$+1$, and similarly for $v$.
We can also follow this method with a greedy method in which we alternatively 
consider swapping
each pair of rows, one from each group, and each pair of columns, one from 
each group.

\clearpage
\section{Extensions and variations}\label{s-extension}
In this section we introduce various extensions for MLR and related methods. 
In particular we present an extension of the rank allocation method, and 
provide generalizations of MLR matrices that include different sparsity patterns and recursion on the block terms.

\subsection{Rank allocation}
In \S\ref{s-rank-alloc} we introduce a method for distributing MLR-rank 
across all levels. This method operates by identifying two distinct levels for
which rank exchange maximizes the predicted net decrease.
More specifically, it is achieved by exchanging one unit of rank between two levels.
The idea can be extended to identify two distinct levels that can swap
from
$1$ to $q$ units of rank to maximize the decrease in
Frobenius norm squared error.
The corresponding decrease, $\delta^+_l$, and increase, $\delta^-_l$, in error can 
be computed directly using
\[
\delta_l^+ = \sum_{k=1}^{p_l} \sum_{j=1}^{q} \sigma_{r_l+j}^2 (R_{l,k}),
\qquad
\delta_l^- = \sum_{k=1}^{p_l} \sum_{j=1}^{q} \sigma_{r_l-j-1}^2 (R_{l,k}).
\]

This strategy offers greater flexibility in re-allocating ranks.
Lastly, we can retain several top candidates for rank exchange and terminate the 
rank allocation algorithm if none of them decreases the fitting error.

\subsection{Sparsity refinement}
In the MLR definition, the partition at each level is 
hierarchical. This implies that the sparsity pattern of each level's 
block diagonal matrix $A^l$ refines the sparsity pattern 
found in the preceding level's block diagonal matrix $A^{l-1}$. 
However, we can relax the refinement constraint and only require $A^l$ to be 
a block diagonal matrix. Given that $A$ decomposes onto a sum of 
matrices $A^l$, the BCD algorithm remains directly 
applicable. Moreover, since $A^l$ maintains its block diagonal 
form, $A$ can also be expressed in factor form, making the ALS 
method equally suitable. It is straightforward to see that 
the rank allocation algorithm can be readily applied 
in this context as well.

\subsection{Generalized multilevel low rank matrix}
We further generalize the definition of MLR as follows. 
Define a tail sum as $\bar A^l = \sum_{l'=l}^L A^{l'}$.
The definition of $A$ in \S\ref{s-contig-MLR} can then 
be alternatively presented as a recursion
\BEQ\label{e-mlr-a-rec}
\bar A^{l} = \blkdiag(\bar A_{l,1}, \ldots, \bar A_{l,p_{l}} ) = 
\blkdiag(B_{l,1} C_{l,1}^T, \ldots, 
B_{l,p_l} C_{l,p_l}^T ) + \bar A^{l+1},
\EEQ
where $B_{l,k} \in \reals^{m_{l,k}\times r_l}$ 
and $C_{l,k} \in \reals^{n_{l,k}\times r_l}$,
$\bar A_{l,k}$ is $m_{l,k}\times n_{l,k}$ block of $\bar A^l$ 
restricted to rows
$\left \{\sum_{k'=1}^{k-1} m_{l,k'}+1, \ldots, \sum_{k'=1}^k m_{l,k'}\right \}$ 
and to columns 
$\left \{\sum_{k'=1}^{k-1} n_{l,k'}+1, \ldots, \sum_{k'=1}^k n_{l,k'} \right\}$.
Using the recursion in \eqref{e-mlr-a-rec}, we can write
\[
A=\bar A^1 = B_{1,1} C_{1,1}^T + \blkdiag(\bar A_{2,1}, \ldots, 
\bar A_{2,p_2} ).
\]

The recursion process in \eqref{e-mlr-a-rec} occurs within the block 
diagonal term.
This process can be further generalized to include a recursion for each block 
derived from row and column partitioning,
\BEQ\label{e-ex-mlr-a-rec}
\bar A^l = \left[ \begin{array}{ccc}
 \bar A_{l,1,1} &  & \bar A_{l,1,p_{l}} \\
 & \ddots &  \\
 \bar A_{l,p_{l},1} &  & \bar A_{l,p_{l},p_{l}}
\end{array}\right] = \left[ \begin{array}{ccc}
  B_{l,1,1} C_{l,1,1}^T &  & B_{l,1,p_l} C_{l,1,p_l}^T \\
 & \ddots &  \\
  B_{l,p_l,1} C_{l,p_l,1}^T &  & B_{l,p_l,p_l} C_{l,p_l,p_l}^T
\end{array}\right] + \bar A^{l+1},
\EEQ   
where $B_{l,k_1,k_2} \in \reals^{m_{l,k_1}\times r_l}$ 
and $C_{l,k_1,k_2} \in \reals^{n_{l,k_2}\times r_l}$,
$\bar A_{l,k_1,k_2}$ is $m_{l,k_1}\times n_{l,k_2}$ block of 
$\bar A^l$ restricted to rows
$\left \{\sum_{k'=1}^{k_1-1} m_{l,k'}+1, \ldots, \sum_{k'=1}^{k_1} m_{l,k'} \right\}$ 
and to columns 
$\left \{\sum_{k'=1}^{k_2-1} n_{l,k'}+1, \ldots, \sum_{k'=1}^{k_2} n_{l,k'}
\right\}$.
Similarly, we have 
\[
A = \bar A^1 = B_{l,1,1} C_{l,1,1}^T  + \left[ \begin{array}{ccc}
 \bar A_{2,1,1} &  & \bar A_{2,1,p_{2}} \\
 & \ddots &  \\
 \bar A_{2,p_{2},1} &  & \bar A_{2,p_2,p_2}
\end{array}\right].
\]
We refer to the format in \eqref{e-ex-mlr-a-rec} as the generalized multilevel low 
rank matrix (GMLR).

Since $A$ decomposes into a sum of block matrices at each level
with each block
of a given rank, the BCD algorithm remains applicable. 
Similarly, $A$ is bilinear in $B_{l,k_1,k_2}$ and $C_{l,k_1,k_2}$, 
and therefore ALS scheme can still be applied to GMLR.
Furthermore, given the total rank and matrix partitioning at each level, 
the rank allocation method directly extends to this setting. 
In particular, estimating the increase $\delta_l^+$ (or decrease  $\delta_l^-$) 
in fitting error 
now involves singular values of all blocks on a given level,
rather than just the block diagonal blocks as in MLR. Hence, 
while the rank allocation method in \S\ref{s-rank-alloc} was developed 
for MLR matrices, it can be readily applied to more general settings, 
such as GMLR.

We can further relax the rank constraint to allow blocks at the same level 
to have different ranks.
Consequently, the GMLR without the rank constraint contains a class 
of hierarchical matrices $\mathcal{H}$ as well as BLR and multilevel BLR.

\clearpage
\section{Numerical examples}\label{s-examples}
In this section we illustrate our methods with several numerical examples.
We report the relative fitting error ${\|A-\hat A\|_F}/{\|A\|_F}$ 
and rank allocation evolution versus the number of iterations.  
We use three different initial allocations of rank,
MLR bottom, MLR uniform, and MLR top, as described in \S\ref{s-rank-alloc}.
We also report the error of the low rank approximation, denoted as LR.
For square matrices we report the low rank plus diagonal 
approximation (denoted as LR+D) or factor model in the symmetric PSD case.

The stopping criterion for
factor fitting~\eqref{e-stopping-crit} is
$\epsilon_{\text{rel}}=0.01$ and for rank allocation,
$\epsilon_{\text{rel}}=0.001$. 
In factor fitting for the low rank plus diagonal model, 
we use $\epsilon_{\text{rel}}=10^{-6}$.  
For each iteration of rank allocation, we 
carry out $2$ V-epochs of BCD during factor fitting. 
In \S\ref{s-ex-asset-cov} we use a given hierarchical partitioning;
in \S\ref{s-rad-kern} the hierarchical partitioning is derived based on
recursive spectral bi-clustering of a spatial domain,
while in the remaining examples, we employ hierarchical partitioning 
found by spectral dissection, as detailed in \S\ref{s-MLR-fitting}.
In the process of constructing a hierarchical partitioning, 
we refine each partitioning of the rows and columns (found by
spectral dissection) using a 
greedy algorithm for at most $5000$ pair swaps that decrease
the objective.

We compare the performance of MLR with HODLR and Monarch matrices.
We report the fitting errors associated with HODLR and Monarch matrices, 
while matching their storage with MLR.
HODLR is specified by the block fitting tolerance and 
hierarchical partitioning.
We conduct a comparison with the HODLR, adjusting its tolerance parameter to 
approximately match the storage of other formats.
In every experiment, we use the same hierarchical partitioning for HODLR 
that we do for the MLR with the best fitting error.
Note that due to the difference in specification of the formats 
(MLR, LR and LR+D are defined by rank, while HODLR is defined by tolerance), 
the comparison with HODLR may not be entirely fair.

We employ the official implementation from~\cite{dao2022monarch} to find 
an analytical solution for fitting a dense matrix using a Monarch matrix. 
We apply a global permutation to the input matrix given by its hierarchical 
partitioning, 
as it results in a smaller fitting error compared to an unpermuted matrix.
We search over several parameters like block sizes and rank, 
and we report the best fitting error among them. 
Nevertheless, due to the difference in format specification with MLR, 
this comparison is not entirely fair.

\subsection{Asset covariance matrix}~\label{s-ex-asset-cov}
\begin{figure}
\begin{center}
\begin{tikzpicture}
\coordinate (A) at (-5,0) {};
\coordinate (B) at ( 5,0) {};
\coordinate (C) at (0,6) {};
\draw[name path=AC] (A) -- (C);
\draw[name path=BC] (B) -- (C);
\foreach \y/\A in {0/$5000$ assets,1/$157$ sub-industries,
                2/$69$ industries,3/$24$ groups,4/$11$ sectors,5/ root} {
    \path[name path=horiz] (A|-0,\y) -- (B|-0,\y);
    \draw[name intersections={of=AC and horiz,by=P},
          name intersections={of=BC and horiz,by=Q}] (P) -- (Q)
        node[midway,above] {\A};
}
\end{tikzpicture}
\end{center}
\caption{GICS hierarchical partition of 5000 companies.}\label{f-gics}
\end{figure}
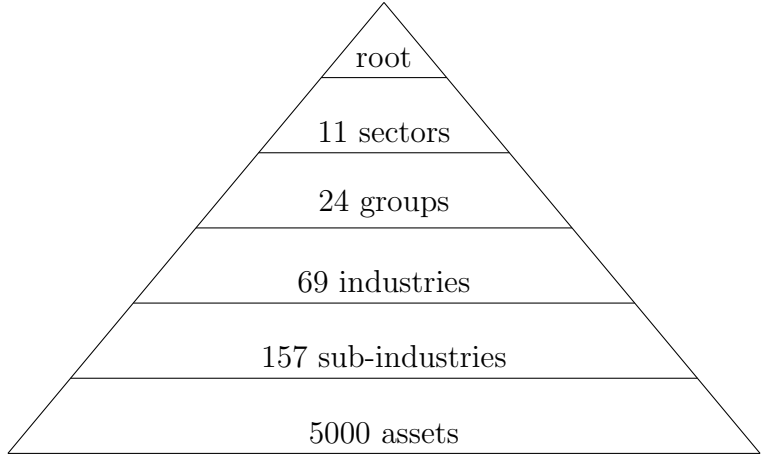

We consider the problem of fitting the empirical covariance matrix of
daily stock returns.
For this example there is a widely used hierarchical partition of the
assets, the Global Industry Classification Standard (GICS)~\cite{bhojraj2003s},
based on the company's primary business activity.
This hierarchical partition has $L=6$ levels, shown in figure~\ref{f-gics}.

In this example the daily returns are found
or derived from data from CRSP Daily Stock and  CRSP/Compustat 
Merged Database
\textcopyright2023 Center for Research in Security Prices 
(CRSP$^\text{\textregistered}$), 
The University of Chicago Booth School of Business. 
We consider a 300 (trading) day period ending 2022/12/30.  
We obtain the
GICS codes from CRSP/Compustat Merged Database 
-- Security Monthly during 2022/06/30 to 2023/01/31. 
We joined the two tables using PERMCO (a unique permanent 
company identification number assigned by CRSP to all companies). 
Next we filtered out assets with missing GICS codes,
and kept the $m=5000$ assets with the fewest missing return values.
(We understand that this introduces survival bias, but our goal here is 
only to illustrate MLR fitting, and not build a covariance model.)
We clip or Winsorize the entries at $\pm 3 \sigma$, where 
$\sigma$ is the standard deviation of return of each stock over 
the period.
We let $\Sigma$ be the $5000 \times 5000$ empirical covariance matrix
of the returns.  Since it is based on only $300$ days, its rank is
$300$.
We fit $\Sigma$ with a PSD MLR matrix of total rank $r=30$.

\begin{figure}
    \begin{center}
    \includegraphics[width=0.7\textwidth]{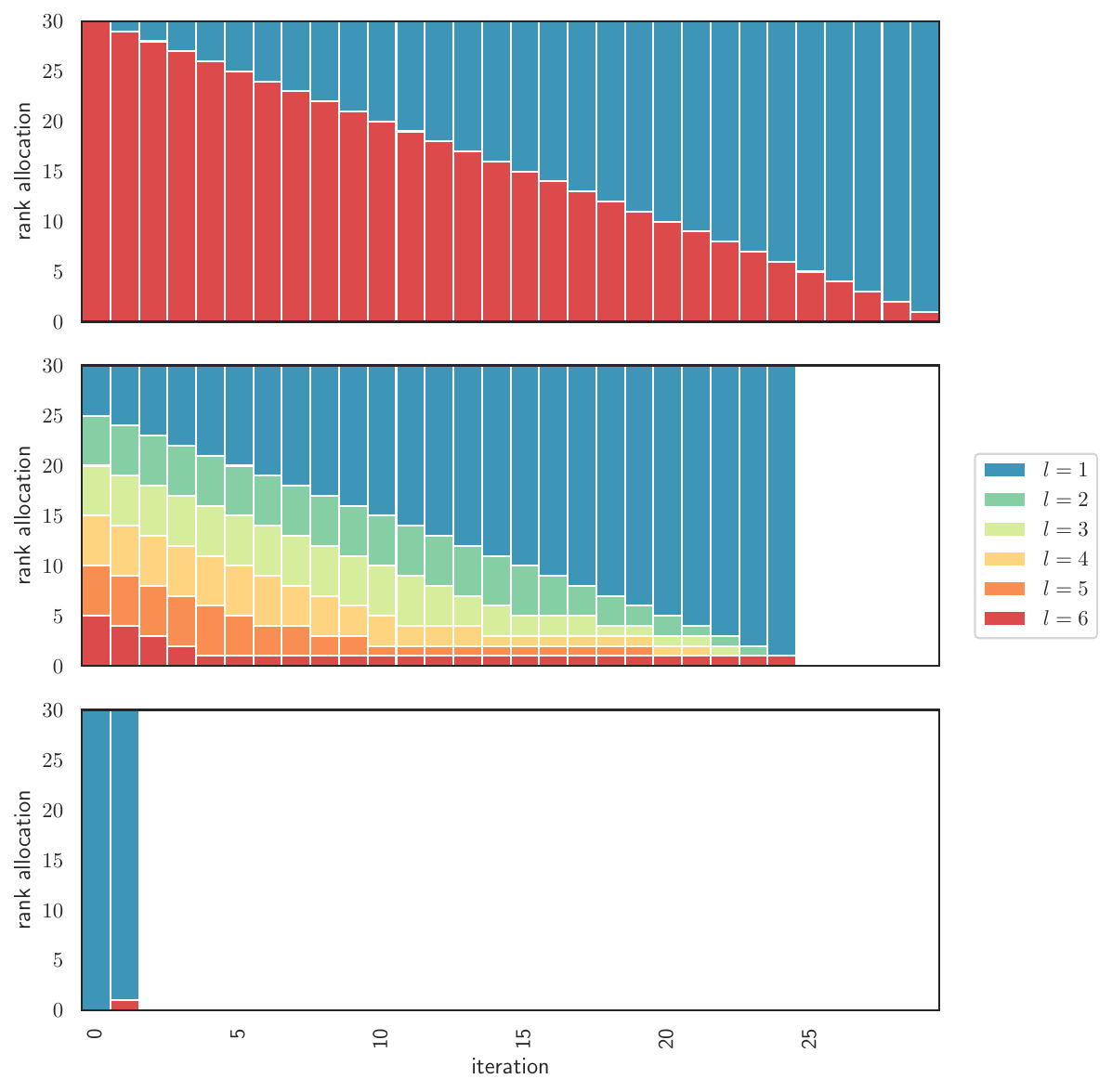}
    \end{center}
    \caption{Rank $r=30$ partitioning across $L=6$ levels 
    during fitting for asset covariance matrix with GICS hierarchy,
starting from the bottom level $l=6$, uniform and the top level $l=1$ initial 
    allocation. }
    \label{fig-cov-ra-evolution}
\end{figure}

\paragraph{Results.}
A factor model is the traditional approach for modeling a covariance 
matrix of financial returns~\cite{coughlin2013analysis};
we were curious to see if a different MLR approximation might give 
a better fit to the empirical covariance matrix.
Using the GICS hierarchical partition, we ran rank allocation 
from $3$ different initial allocations 
(see table~\ref{tab-cov-error}). 
In all three cases the method converged to the factor model, \ie,
$r_1=29,~r_2=\cdots=r_5=0$ and $r_6=1$
(see figure~\ref{fig-cov-ra-evolution}).
In particular, the GICS hierarchy appears to bring no advantage
over the standard factor model,
at least in terms of the Frobenius norm fitting of the 
empirical covariance matrix.
(Still, the GICS classification is widely used
in portfolio construction~\cite{bhojraj2003s,
briere2016factor} and in finding stable 
eigenportfolios~\cite{avellaneda2020hierarchical, serur2020hierarchical}.)


\begin{table}
\centering
\begin{tabular}{l  l l} 
 Method & Error $(\%)$ & Storage $(\times 10^5)$ \\  
 \hline
  LR &  $16.178$ & $1.500$\\
  LR+D &  $\bf{15.379}$ & $1.500$\\
  HODLR &  $38.837$ & $1.503$\\
  Monarch &  $17.971$ & $1.560$\\
  MLR bottom &  $\bf{15.379}$ & $1.500$\\
  MLR uniform &  $\bf{15.379}$ & $1.500$\\
  MLR top &  $\bf{15.379}$ & $1.500$\\
\end{tabular}
\caption{Fitting errors for asset covariance matrix with GICS hierarchical 
partition.}
\label{tab-cov-error}
\end{table}

\begin{figure}
    \begin{center}
    \includegraphics[width=0.7\textwidth]{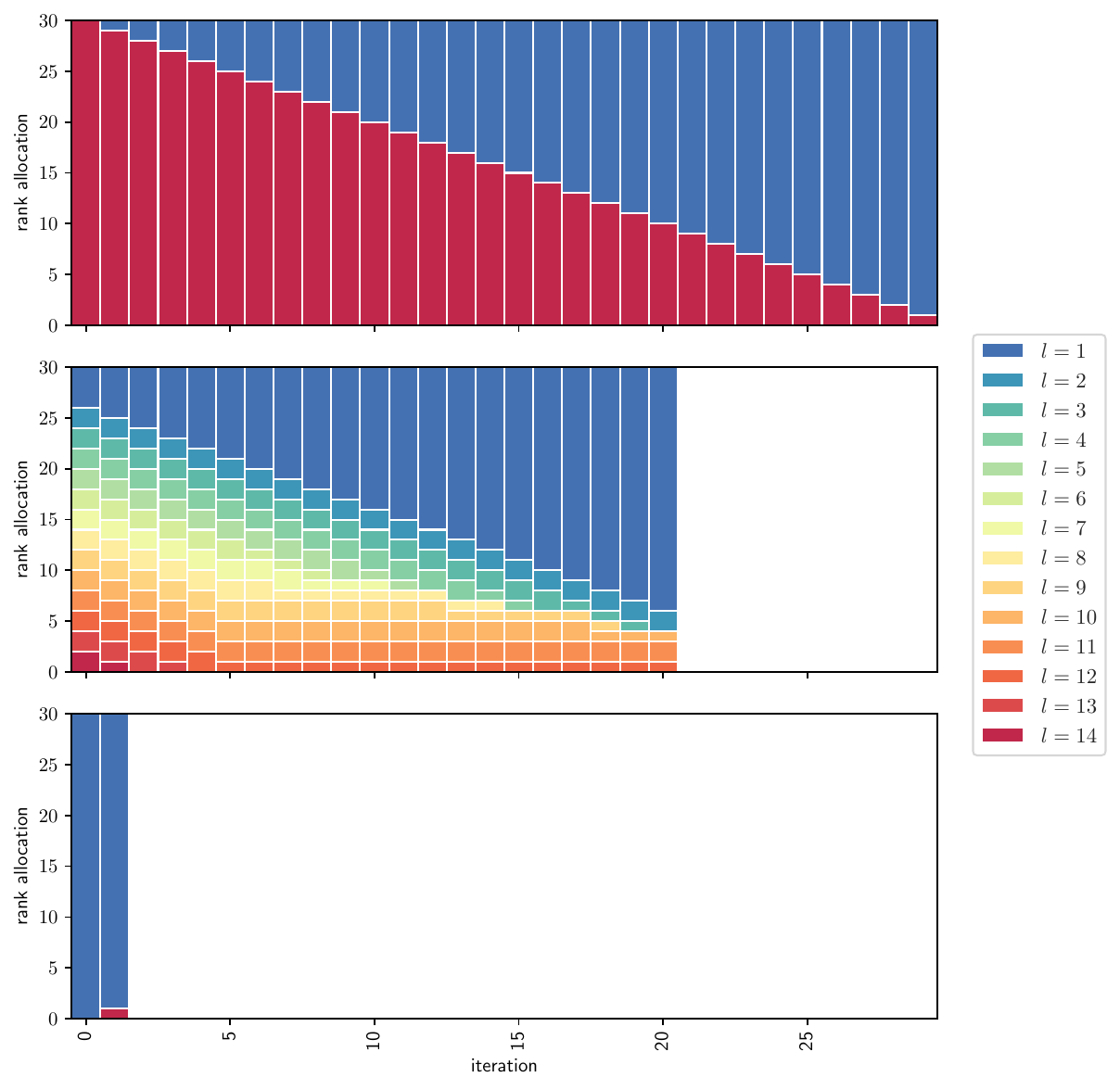}
    \end{center}
    \caption{Rank $r=30$ partitioning across $L=14$ levels 
    during fitting for asset covariance matrix with full hierarchy, starting 
    from the bottom level $l=14$, uniform ($r_{1}=4$, $r_2 = \cdots =r_{14}=2$) 
    and the top level $l=1$ initial allocation.}
    \label{fig-cov-ra-evolution-full}
\end{figure}

We also ran our full MLR fitting method, ignoring the GICS hierarchy,
fitting an MLR matrix with $L=14$ levels.
For two out of the three initial rank allocations, our method converged
again to the factor model;
see table~\ref{tab-cov-error-full} and 
figure~\ref{fig-cov-ra-evolution-full}.
We take these results as an endorsement of the commonly used 
factor model.

The fitting error achieved by HODLR is at minimum twice as large as that of 
low rank approximation across both hierarchical partitionings.
The Monarch fitting error is comparable but worse than that of the LR.


\begin{table}
\centering
\begin{tabular}{l l l} 
 Method & Error $(\%)$  & Storage $(\times 10^5)$\\  
 \hline
  LR &  $16.178$ & $1.500$\\
  LR+D &  $\bf{15.379}$ & $1.500$\\
  HODLR &  $34.643$ & $1.509$\\
  Monarch &  $17.980$ & $1.560$\\
  MLR bottom &  $\bf{15.379}$ & $1.500$\\
  MLR uniform &  $15.685$& $1.500$\\
  MLR top &  $\bf{15.379}$& $1.500$\\
\end{tabular}
\caption{Fitting errors for asset covariance matrix with  full hierarchy.}
\label{tab-cov-error-full}
\end{table}

\clearpage
\subsection{Fiedler matrix}\label{s-fiedlerl}
Here we consider a Fiedler matrix, with entries
$A_{ij} = |a_i - a_j|$ for some vector $a\in\reals^n$. 
We set $m=n=5000$ and sample the entries of $a$ 
uniformly from $[0,1]$. 
We fix the rank as $r=28$, and apply our fitting method
to find symmetric (but not PSD) MLR matrix approximations.

\begin{figure}
    \begin{center}
    \includegraphics[width=0.7\textwidth]{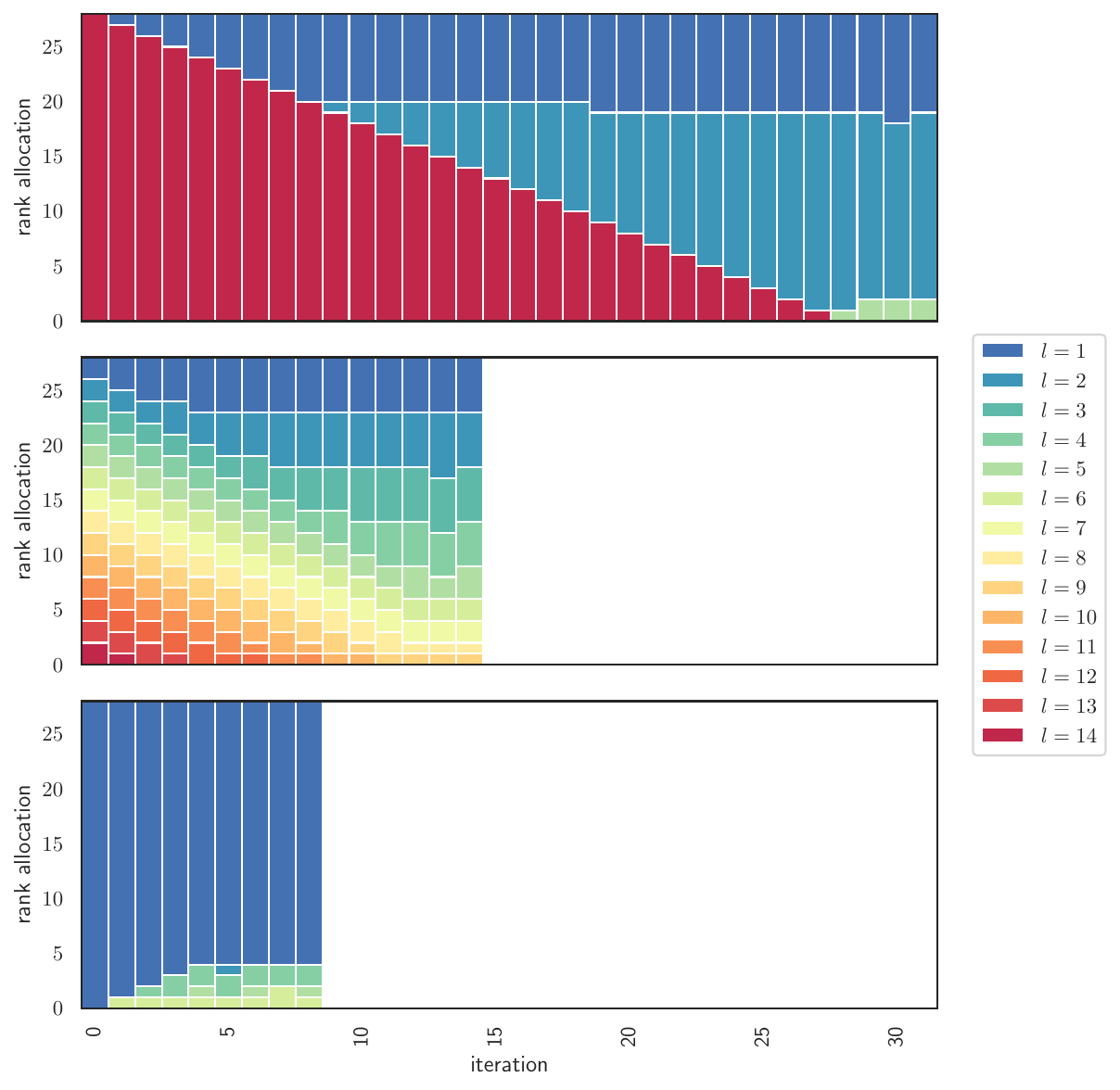}
    \end{center}
    \caption{Rank $r=28$ partitioning across $L=14$ levels 
    during fitting of the Fiedler matrix, starting from the bottom level $l=14$, 
    uniform and the top level $l=1$ initial allocation. }
    \label{fig-fielder-ra-evolution}
\end{figure}

\paragraph{Results.}
The algorithm converges to three different MLR matrices, all better than
a low rank approximation.  The MLR matrix found when initialized with 
a uniform allocation has objective
that is almost $20\times$ smaller than the low rank model;
see table~\ref{tab-fiedler-error} and
figure~\ref{fig-fielder-ra-evolution}.
Its rank allocation is
\[
\begin{array}{c}
r_1=r_2=r_3=5,\quad r_4=4, \quad r_5=3,\\ r_6=r_7=2, \quad r_8=r_9=1,
\quad r_{10}=\cdots=r_{14}=0.
\end{array}
\]
HODLR achieves the same fitting error as MLR with the uniform 
initialization while having smaller storage.
The error achieved by the Monarch matrix is lower than the LR by $4\%$.



\begin{table}
\centering
\begin{tabular}{l l l} 
 Method & Error $(\%)$  & Storage $(\times 10^5)$\\  
 \hline
  LR &  $0.1988$ & $1.400$\\
  LR+D &  $0.2081$ & $1.400$\\
  HODLR &  $\bf{0.0128}$ & $1.373$\\
  Monarch &  $0.1903$ & $1.500$\\
  MLR bottom &  $0.0996$ & $1.400$\\
  MLR uniform &  $\bf{0.0128}$ & $1.400$\\
  MLR top &  $0.0548$ & $1.400$\\
\end{tabular}
\caption{Fitting errors for Fielder matrix.}
\label{tab-fiedler-error}
\end{table}

\clearpage
\subsection{Discrete Gauss transform matrix}\label{s-dgt}
The discrete Gauss transform (DGT) matrix~\cite{greengard1991fast} 
is given by
\[
A_{ij} = e^{-\|t_i-s_j\|_2^2/h^2},
\]
where $s_j\in\reals^d$ for $j=1,\ldots, n$ and $t_i\in\reals^d$ for $i=1,\ldots, m$
are source and target locations respectively and $h>0$ is the bandwidth.
Similarly to the experimental setup in~\cite{yang2003improved}, we set 
sources and targets to be uniformly distributed in a unit cube $[0,1]^d$ 
with $d=3$, $m=5000$, $n=7000$, and $h=0.2$. We fix the rank as $r=28$.

\begin{figure}
    \begin{center}
    \includegraphics[width=0.7\textwidth]{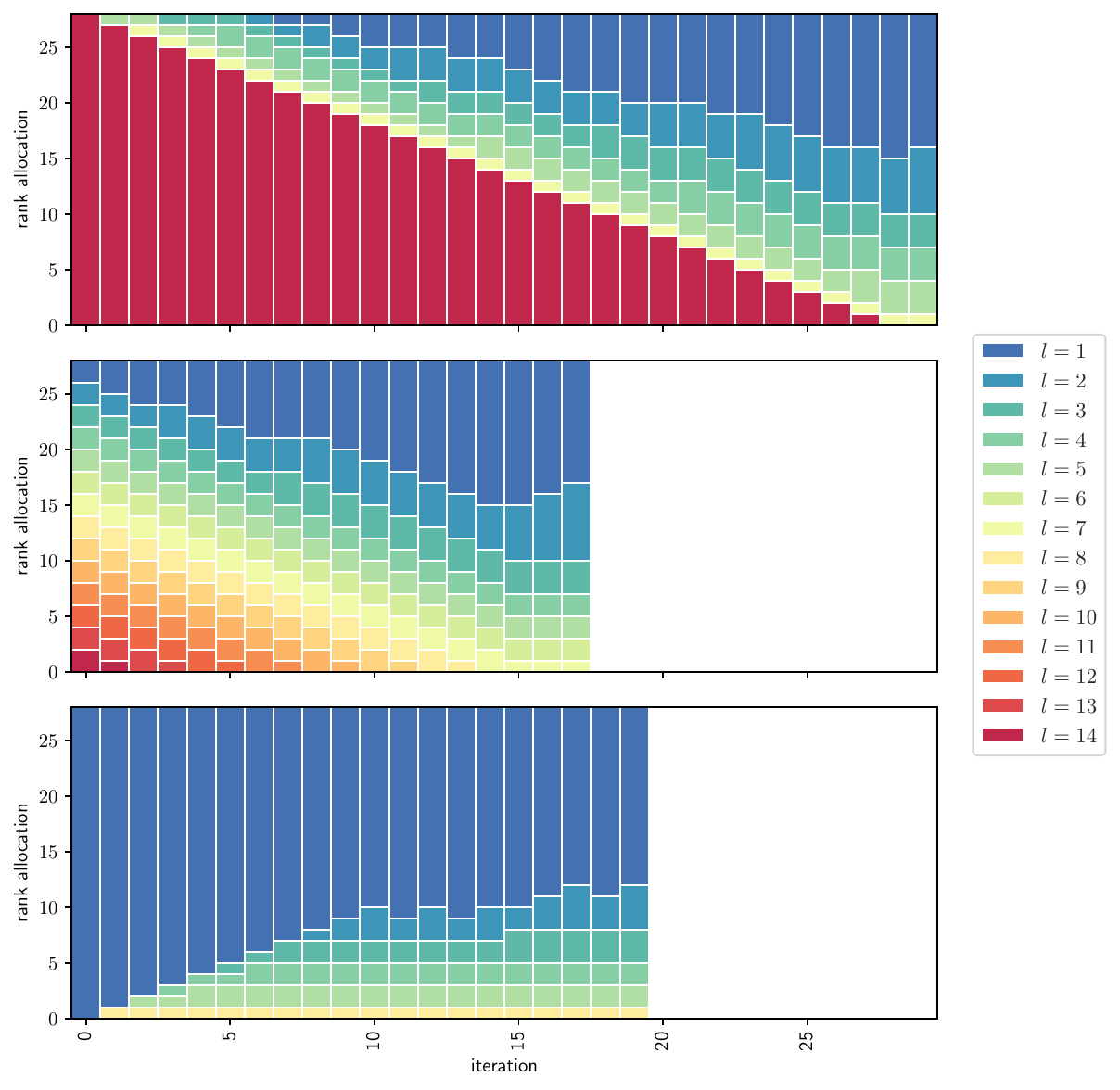}
    \end{center}
    \caption{Rank $r=28$ partitioning across $L=14$ levels 
    during fitting of DGT matrix, starting from the bottom level $l=14$, 
    uniform and the top level $l=1$ initial allocation.  }
    \label{fig-dgt-ra-evolution}
\end{figure}

\paragraph{Results.}
Our algorithm converges to three different MLR matrices,
each better than the low rank or low rank plus diagonal approximations or HODLR
or Monarch;
see table~\ref{tab-dgt-error}
and figure~\ref{fig-dgt-ra-evolution}.
The lowest approximation error is achieved with
the bottom rank initialization, and achieves error more
than a factor $2$ smaller than the low rank model.
It has rank allocation 
\[
\begin{array}{c}
r_1=12,\quad r_2=6, \quad r_3=r_4=r_5=3, \quad r_6=0, 
\quad r_7=1, 
\quad r_8=\cdots=r_{14}=0.
\end{array}
\]            
 The HODLR and Monarch objectives surpass that of the MLR by more than a 
 factor of $4$ and $2$ respectively.



\begin{table}
\centering
\begin{tabular}{l  l l} 
 Method & Error $(\%)$ & Storage $(\times 10^5)$ \\  
 \hline
  LR &  $41.779$ & $3.360$\\
  HODLR & $72.549$ & $3.385$ \\
  Monarch & $43.962$ & $3.600$ \\
  MLR bottom &  $\bf{16.753}$ & $3.360$\\
  MLR uniform &  $21.766$ & $3.360$\\
  MLR top &  $25.759$ & $3.360$\\
\end{tabular}
\caption{Fitting errors for DGT matrix.}
\label{tab-dgt-error}
\end{table}

\clearpage
\subsection{Distance matrix}\label{sec-ex-spdist}
Consider a connected weighted graph $G=(V,E,w)$ 
with $V=\{1,2,\ldots,n\}$ nodes and $E=\{1,2,\ldots,d\}$ edges. 
Let $D \in \reals_+^{n \times n}$ denote the
shortest path distance matrix, \ie,  
$D_{ij}$ is a shortest path distance from node $j$ to node $i$. 
We will approximate $D$ by a symmetric MLR matrix.

We consider the drivable street network of Venice, Italy. 
It contains  $n=5893$ nodes and $12026$ edges.
We use the OSMnx Python package~\cite{boeing2017osmnx} to access the data
in OpenStreetMap~\cite{OpenStreetMap}.
This graph is directed; to obtain an undirected graph,
we ignore the direction and take the weight to be the average of
the forward and reverse directions.
This results in a connected graph with $12098$ nonzero entries in
the adjacency matrix.
We use our MLR fitting method to obtain MLR matrices with
$L=14$ levels, with a total rank to be allocated of $r=98$.

\begin{figure}
    \begin{center}
    \includegraphics[width=0.7\textwidth]{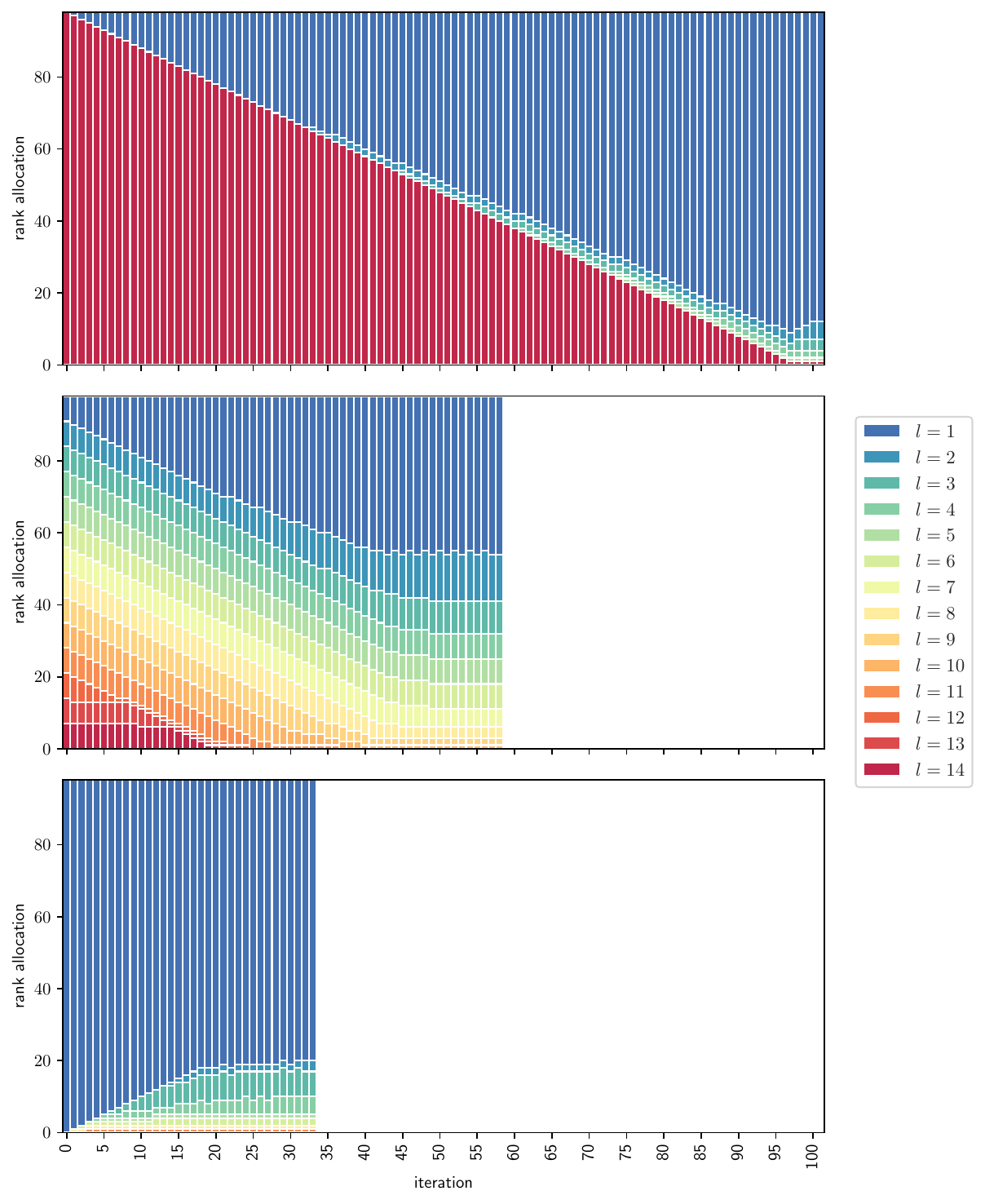}
    \end{center}
    \caption{Rank $r=98$ partitioning across $L=14$ levels 
    during fitting of the distance matrix, starting from the bottom level $l=14$, 
    uniform and the top level $l=1$ initial allocation. }
    \label{fig-dist-ra-evolution}
\end{figure}

\paragraph{Results.}
Table~\ref{tab-dist-error} shows the error obtained from
three initial rank allocations.  All are better 
than the low rank approximation, with the one obtained from
a uniform allocation obtaining an error about $2\times$ smaller.
The lowest fitting error is attained with uniform allocation, where
\[
\begin{array}{c}
r_1=44, \quad r_2=13,\quad  r_3=9, \quad r_4=r_5=r_6=7,\\
r_7=5, \quad  r_8=3, \quad r_9=2, \quad
r_{10}=1, \quad r_{11}=\cdots=r_{14}=0.
\end{array}
\]
The rank allocations versus iterations are shown in
figure~\ref{fig-dist-ra-evolution}.
The fitting error of MLR is more than $6\times$ less
than the HODLR approximation error
and $2\times$ less than the Monarch 
approximation error.


\begin{table}
\centering
\begin{tabular}{l l l} 
 Method & Error $(\%)$ & Storage $(\times 10^5)$\\  
 \hline
  LR &       $0.7197$ & $5.775$ \\
  LR+D &     $0.7056$ & $5.775$ \\
  HODLR &     $2.4696$ & $5.789$ \\
  Monarch &     $0.8692$ & $5.880$ \\
  MLR bottom &  $0.6644$ & $5.775$ \\
  MLR uniform &  $\bf{0.3702}$ & $5.775$ \\
  MLR top &  $0.3929$ & $5.775$ \\
\end{tabular}
\caption{Fitting errors for distance matrix.}
\label{tab-dist-error}
\end{table}

\clearpage
\subsection{Multiscale inverse polynomial kernel matrix}\label{s-rad-kern}
The multiscale inverse polynomial kernel matrix 
is given by
\[
A_{ij} = \sum_{l=0}^{L_A-1} \left(1+\left(\frac{ \|t_i-s_j\|_2}{\sigma / 2^l}
\right)^2\right)^{-2},
\]
where $s_j\in\reals^d$ for $j=1,\ldots, n$ and $t_i\in\reals^d$ for 
$i=1,\ldots, m$
are source and target locations respectively. To induce different scaling, 
we decrease the kernel radius $\sigma/2^l$ with every level increment 
$l=0, \ldots, L_A-1$.
We set 
sources and targets to be uniformly distributed on a unit sphere in $\reals^{d}$ 
with $d=3$, $m=n=5000$, $L_A=3$, and $\sigma=0.9$.
We derive the hierarchical partitioning based on the recursive spectral 
bi-clustering of source and target points w.r.t. the Euclidean distances 
$\|t_i-s_j\|_2$, and use it for all MLR, HODLR and Monarch. 
We fix the rank as $r=28$.

\begin{figure}
    \begin{center}
    \includegraphics[width=0.7\textwidth]{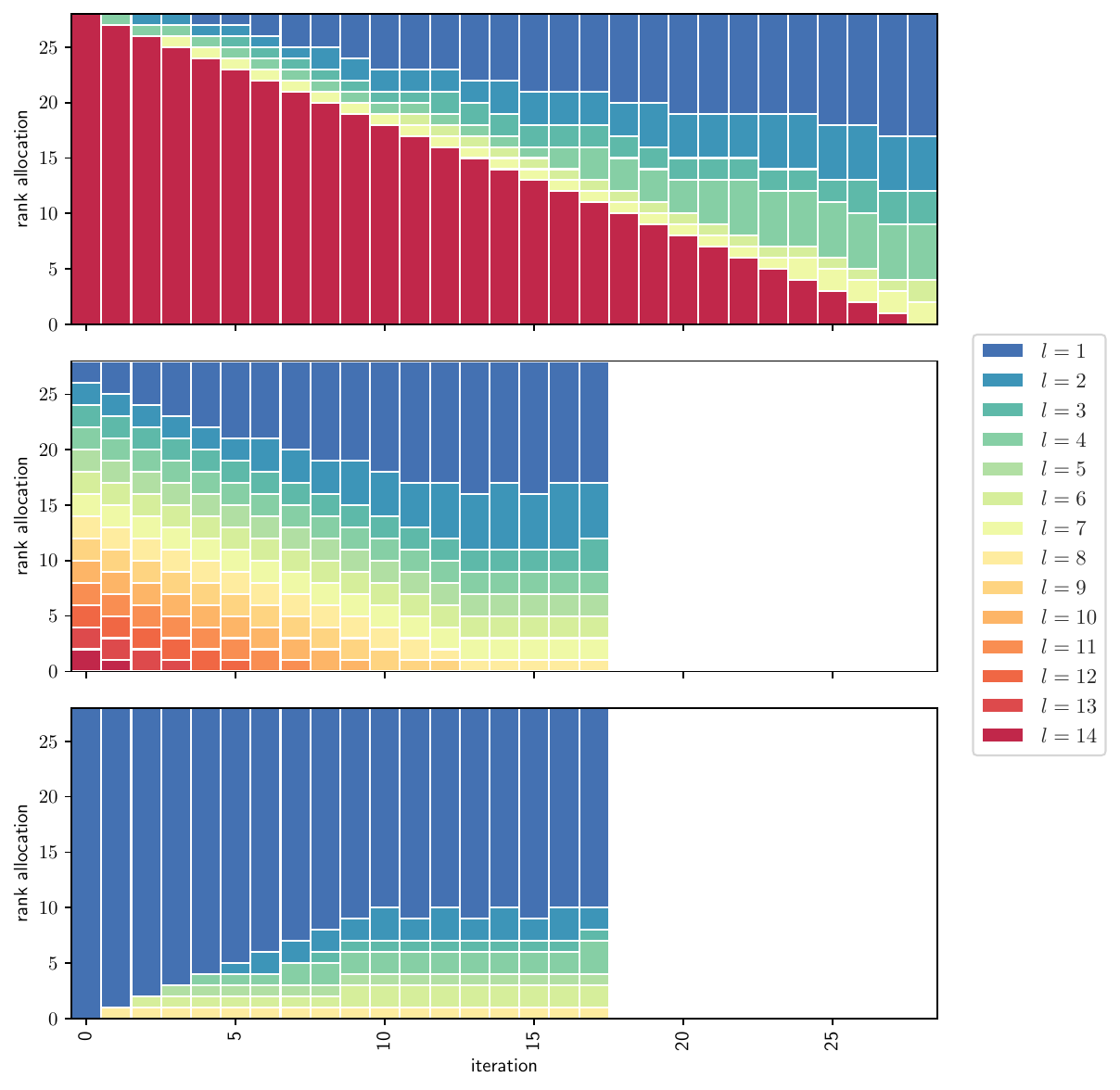}
    \end{center}
    \caption{Rank $r=28$ partitioning across $L=14$ levels 
    during the fitting of multiscale inverse polynomial kernel matrix, 
    starting from the bottom level $l=14$, 
    uniform and the top level $l=1$ initial allocation.  }
    \label{fig-rad-kern-ra-evolution}
\end{figure}

\paragraph{Results.}
The algorithm converges to three different MLR matrices,
each better than the low rank or HODLR approximations;
see table~\ref{tab-rad-kern-error}
and figure~\ref{fig-rad-kern-ra-evolution}.
The lowest approximation error is achieved with
the bottom rank initialization, with error more
than a factor $3$ smaller than the low rank model.
It has rank allocation 
\[
\begin{array}{c}
r_1=11,\quad r_2=5, \quad r_3=3, \quad r_4=5, \\
\quad r_5=0, \quad r_6=r_7=2, \quad r_8=\cdots=r_{14}=0.
\end{array}
\]
The approximation errors of the HODLR and Monarch are respectively $6\times$ and 
$4\times$ greater than the fitting error produced by the MLR.


\begin{table}
\centering
\begin{tabular}{l  l l} 
 Method & Error $(\%)$ & Storage $(\times 10^5)$ \\  
 \hline
  LR &  $22.767$ & $2.800$\\
  LR+D &  $23.282$ & $2.800$\\
  HODLR & $40.657$ & $2.813$ \\
  Monarch & $25.682$ & $2.800$ \\
  MLR bottom &\hskip 5pt $\bf{6.497}$ & $2.800$\\
  MLR uniform &\hskip 6pt  $7.733$ & $2.800$\\
  MLR top & \hskip 6pt $9.400$ & $2.800$\\
\end{tabular}
\caption{Fitting errors for multiscale inverse polynomial kernel matrix.}
\label{tab-rad-kern-error}
\end{table}

\subsection*{Acknowledgments}
The authors thank Parth Nobel, Mykel Kochenderfer,
Emmanuel Candes, and Dylan Rueter
for useful discussions and helpful suggestions.
We thank Liu Kan for pointing out a minor error in an early version.
We gratefully acknowledge support from the Oliger Memorial Fellowship.

\clearpage
\bibliography{references}
\clearpage

\end{document}